\documentclass[letterpaper, 10 pt, conference]{ieeeconf}  

\IEEEoverridecommandlockouts                              

\makeatletter
\let\NAT@parse\undefined
\makeatother
\usepackage[colorlinks,
            linkcolor=blue,
            anchorcolor=blue,
            citecolor=blue]{hyperref}

\usepackage{graphics} 
\usepackage{amssymb}  
\usepackage{amsmath,amsfonts}
\usepackage{algorithmic}
\usepackage{algorithm}
\usepackage{array}
\usepackage{textcomp}
\usepackage{url}
\usepackage{verbatim}
\usepackage{graphicx}
\usepackage{cite}
\usepackage{subfigure}
\usepackage{booktabs}
\usepackage{multirow}
\usepackage{multicol}
\usepackage{makecell}

\title{\LARGE \bf
Autonomous Multiple-Trolley Collection System with Nonholonomic Robots: Design, Control, and Implementation
}

\author{$\text{Peijia~Xie}^{\dagger}$,
        $\text{Bingyi~Xia}^{\dagger}$, 
        Anjun~Hu, 
        Ziqi~Zhao,
        Lingxiao~Meng,\\
        Zhirui~Sun,
        Xuheng~Gao,
        Jiankun~Wang, 
        and Max~Q.-H.~Meng
        \thanks{$^{\dagger}\text{Equal contribution.}$}
        \thanks{This work is partially supported by National Natural Science Foundation of China grant \#62103181. \emph{(Corresponding authors: Bingyi Xia.)}}
        \thanks{Peijia Xie, Bingyi Xia, Anjun Hu, Ziqi Zhao, Lingxiao Meng, Zhirui Sun, Xuheng Gao,  Jiankun Wang and Max Q.-H. Meng are with Shenzhen Key Laboratory of Robotics Perception and Intelligence, Department of Electronic and Electrical Engineering, Southern University of Science and Technology, Shenzhen, China.
        {\tt \small\{xiepj2022, xiaby2020, 12332158, zhaozq2020, menglx2021, sunzr2023, gaoxh2021\}@mail.sustech.edu.cn, wangjk@sustech.edu.cn, max.meng@ieee.org}}
        \thanks{Jiankun Wang is also with the Jiaxing Research Institute, Southern University of Science and Technology, Jiaxing, China.}
        \thanks{Max~Q.-H.~Meng is also a Professor Emeritus in the Department of Electronic Engineering at the Chinese University of Hong Kong in Hong Kong and was a Professor in the Department of Electrical and Computer Engineering at the University of Alberta in Canada.}
}

\begin{document}

 \maketitle
\thispagestyle{empty}
\pagestyle{empty}

\newcommand{\figref}[1]{Fig.~\ref{#1}}
\newcommand{\mvc}[1]{\mathbf{#1}}
\newcommand{\mvcd}[1]{\dot{\mathbf{#1}}}
\newcommand{\mvcdd}[1]{\ddot{\mathbf{#1}}}
\newcommand{\mfra}[1]{\{#1\}}

\begin{abstract}
The intricate and multi-stage task in dynamic public spaces like luggage trolley collection in airports presents both a promising opportunity and an ongoing challenge for automated service robots.
Previous research has primarily focused on handling a single trolley or individual functional components, creating a gap in providing cost-effective and efficient solutions for practical scenarios.
In this paper, we propose a mobile manipulation robot incorporated with an autonomy framework for the collection and transportation of multiple trolleys that can significantly enhance operational efficiency.
We address the key challenges in the trolley collection problem through the novel design of the mechanical system and the vision-based control strategy.
We design a lightweight manipulator and docking mechanism, optimized for the sequential stacking and transportation of multiple trolleys.
Additionally, based on the Control Lyapunov Function and Control Barrier Function, we propose a novel vision-based control with the online Quadratic Programming which significantly improves the accuracy and efficiency of the collection process.
The practical application of our system is demonstrated in real-world scenarios, where it successfully executes multiple-trolley collection tasks. 

\end{abstract}


\section{Introduction}


Service robots have been developed with autonomous movement and object manipulation capabilities, enabling them to perform tasks while navigating dynamic environments and interacting with humans\cite{Hai2020,WANG2021100001,teng2023fusionplanner,Choi2019}.
Researchers are confronted with challenges on the one hand, such as uncertainty from dynamic environments\cite{ye2023robot}, complex contact forms of interaction\cite{complex2022}, and real-time computation with limited cost\cite{lowcost2021}.
Large-scale public spaces, such as airports, have begun to utilize service robots for laborious tasks like guidance and assistance\cite{large-scale2021}.
One of the practical problems encountered concerns the collection and transportation of luggage trolleys at airports because passengers often deposit luggage trolleys at scattered locations in the airport.
To gather them in designated areas for reuse, fully automated robotic systems can significantly reduce labor costs, enhance workplace safety, and ensure a better user experience by efficiently handling trolley-related tasks. 

\begin{figure}[tb]
    \centering
    \subfigure[]
    {\includegraphics[width=1.0\linewidth]{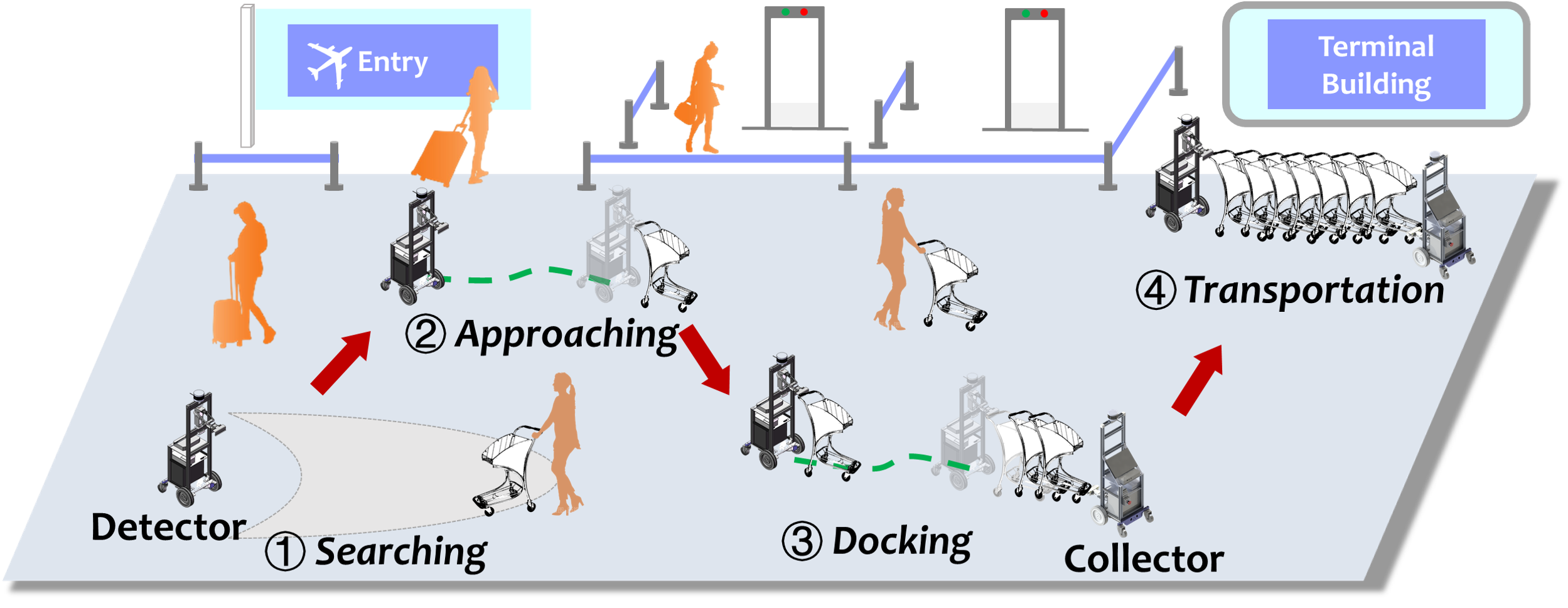}}\\
    \subfigure[]
    {\includegraphics[width=0.9\linewidth]{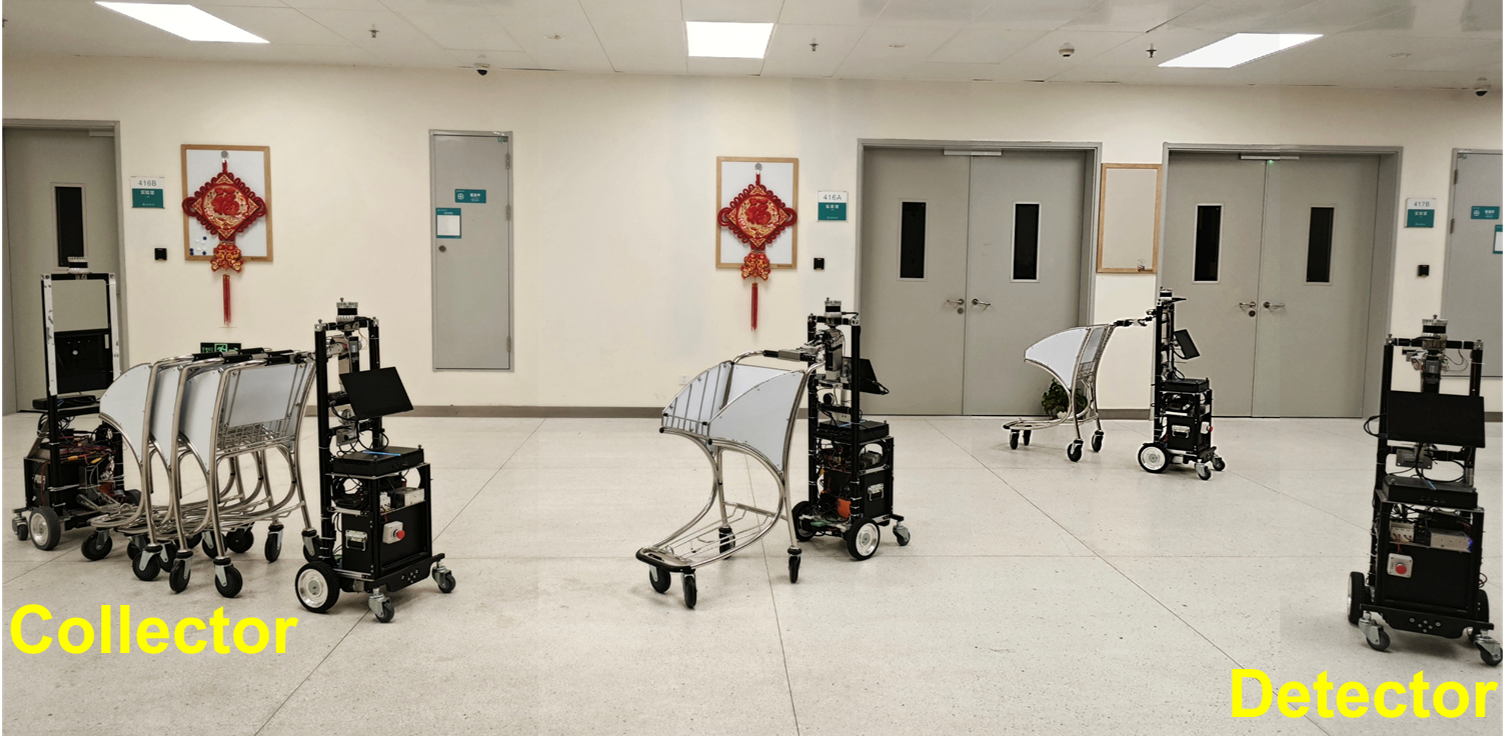}}
    
    \caption{
    (a) Autonomous robots collect and transport luggage trolleys in the airport. 
    The whole task is divided into 4 stages including \textit{Searching, Approaching, Docking, and Transportation} stages respectively. 
    The two robots collaboratively achieve the whole pipeline.
    (b) Demonstration of the proposed robots in the \textit{Approaching and Docking} stages.
    }
    \label{fig:background}
\end{figure}

Several studies present how to develop a robot that executes the task of luggage trolley collection and transportation in the same manner as the operations performed by airport staff.
For the robot catching a trolley and navigating through crowded environments, previous work presents the safety-critical motion planning algorithm\cite{xiao2022robotic} and a decision-making approach for the collection sequence of trolleys\cite{wang2022DecisionMaking}.
Although they are valid, the most efficient deployment operations are to stack multiple luggage trolleys into a queue and transport them simultaneously due to the inherent stackability of luggage trolleys to be stackable when aligning their front and rear ends.
\cite{xia2023collaborative} leverages the property and demonstrates the efficacy of a robotic trolley transportation system where two robots collaboratively steer the head and tail of a trolley queue.
However, it only applies to the already stacked luggage trolley queue in the transportation task.

In this work, we develop an autonomous multi-trolley collection system to address the sequentially stacking task which is defined as the Trolley Collection Problem (TCP).
As shown in \figref{fig:background}, the two nonholonomic robots are referred to as the Detector and the Collector robot respectively. 
The Collector maintains stationary leading the queue of trolleys, while the Detector is responsible for actively searching and manipulating deposited trolleys into alignment.
The robot system works in a sequential navigation and manipulation manner that decomposes the TCP into several functional stages. 
In the searching stage, the Detector robot identifies idle luggage trolleys and locates their positions through onboard sensors. 
Then, the Detector robot arranges multiple luggage trolleys tightly into a queue through mobile manipulation in succession.

Based on the prior studies\cite{xiao2022robotic, wang2022DecisionMaking, xia2023collaborative, pan2021} of the manipulator design paradigm, obstacle avoidance algorithm, and object localization algorithm, we focus on the docking mechanism and the control accuracy in the multi-trolley collection.
First, we develop a practical docking mechanism with low-cost drivers and less freedom for the lightweight robot.
It is designed in consideration of the trade-off between system cost and the number of redundant degrees of freedom (DoF) needed for highly flexible movements. Consequently, the lightweight manipulator implementation limits its capacity to adjust the contact point and the trolley's pose during navigation.
Therefore, the key challenge lies in precisely steering the nonholonomic robot to reach the goal pose estimated by onboard sensors, which is formulated as the same control problem in both the Approaching and Docking stages.
We introduce the Control Lyapunov Function (CLF) and Control Barrier Function (CBF) to the Quadratic Programming (QP), leading to the development of a vision-based controller.

\subsection{Contributions}
The main contributions are summarized as follows:
\begin{itemize}
  \item [1)]
  We develop a robot system with an integrated autonomy framework for the multiple trolley collection task that includes cost-effective hardware design, robust perception, dynamic motion planning, and optimization-base control.
  \item [2)]
  We propose a cost-effective lightweight manipulator and the corresponding docking mechanism for stacking trolleys with limited DoF. 
  \item [3)]
  A vision-based control method is proposed to ensure the Field-of-View (FoV) constraint and high tracking accuracy through CLF-CBF-QP. 
  Experiments show that the proposed controller improves the success rate of the multiple-trolley collection task.
  \item [4)]
  The first demonstration that the low cost of beyond \$10000 robot prototype\footnote{The approximate cost of the robot prototype components are listed in Table~\ref{tab:cost_table}.} safely and stably collects multiple luggage trolleys in a real-world scenario.
\end{itemize}

\subsection{Outline}
Section~\ref{sec:RW} reviews the previous works in visual perception and control for robot mobile manipulation of car-like objects. 
Section~\ref{sec:SD} describes the hardware system and the integrated framework for mobile manipulation. 
Section~\ref{sec:CTR} introduces the CLF-CBF-QP theory and the robot model in the collection tasks and proposes the vision-based control method. 
Section~\ref{sec:VT} introduces our relative measurement method of the target object with different sensors.
Section~\ref{sec:ER} presents the quantitative results of the comparative experiments and demonstrates the whole system in the real world. 
Finally, Section~\ref{sec:CON} is the conclusion.

\section{Related Work}
\label{sec:RW}
\subsection{Mobile Manipulation}
In recent years, there has been a significant proliferation of achievements in the generalization of universal mobile manipulation platforms to encompass versatile robot manipulation and collaboration tasks\cite{honerkamp2023n, Jauhri2022}.
However, in practice, task-oriented robotic systems play a substantial role in both industrial and service applications\cite{yang2020human, kashiri2019centauro, vstibinger2021mobile, sustarevas2022autonomous} due to their reliability and robustness when performing repetitive tasks.
For large and heavy object transportation tasks, employing wheeled robots is a basic and common autonomous solution.
In such cases, when the robot lacks the capability to grasp or manipulate the object arbitrarily, the object is subject to unilateral effects imposed by both the robot and the environment.
The pushing manipulation through a mobile robot is intuitively simple but overcomes such physical limits.
For example, \cite{Bertoncelli2020, Tang2023} demonstrate stable nonprehensile object manipulation on mobile robots, even in cases where a traditional manipulator is not equipped. 

\subsection{Pushing a Car-like Object}
In the context of the TCP, trolleys can be classified as the car-like object whose beams can provide a stable contact point for robot grasping.
Previous works\cite{scholz2011cart, schulze2023trajectory} explore the approach of pushing-car tasks by two-arm mobile robots.
They focus on the safety navigation planning and simplify the stabilization control by fully constraining all degrees of freedom.
The robot equipped with one robot arm provides less redundancy but lower cost.
The constraints of a single contact point can be tackled through global path planning\cite{burget2016bi} and system parameter description\cite{Aguilera2023}.
Compared with these universal robot arms, our previous work\cite{xiao2022robotic} addresses the planning problem with a lightweight manipulator specifically designed for the trolley.
The fundamental assumption of these works is the car-like object is tightly held at the beginning.
However, considering the trolley deployment task, the specific destination for the TCP is the end of an arranged queue of trolleys. 
Therefore, this work focuses on improving the control accuracy of both the approaching and docking stages.

\begin{figure*}[tb!]
    \centering
     \subfigure[]
    {\includegraphics[width=0.58\linewidth]{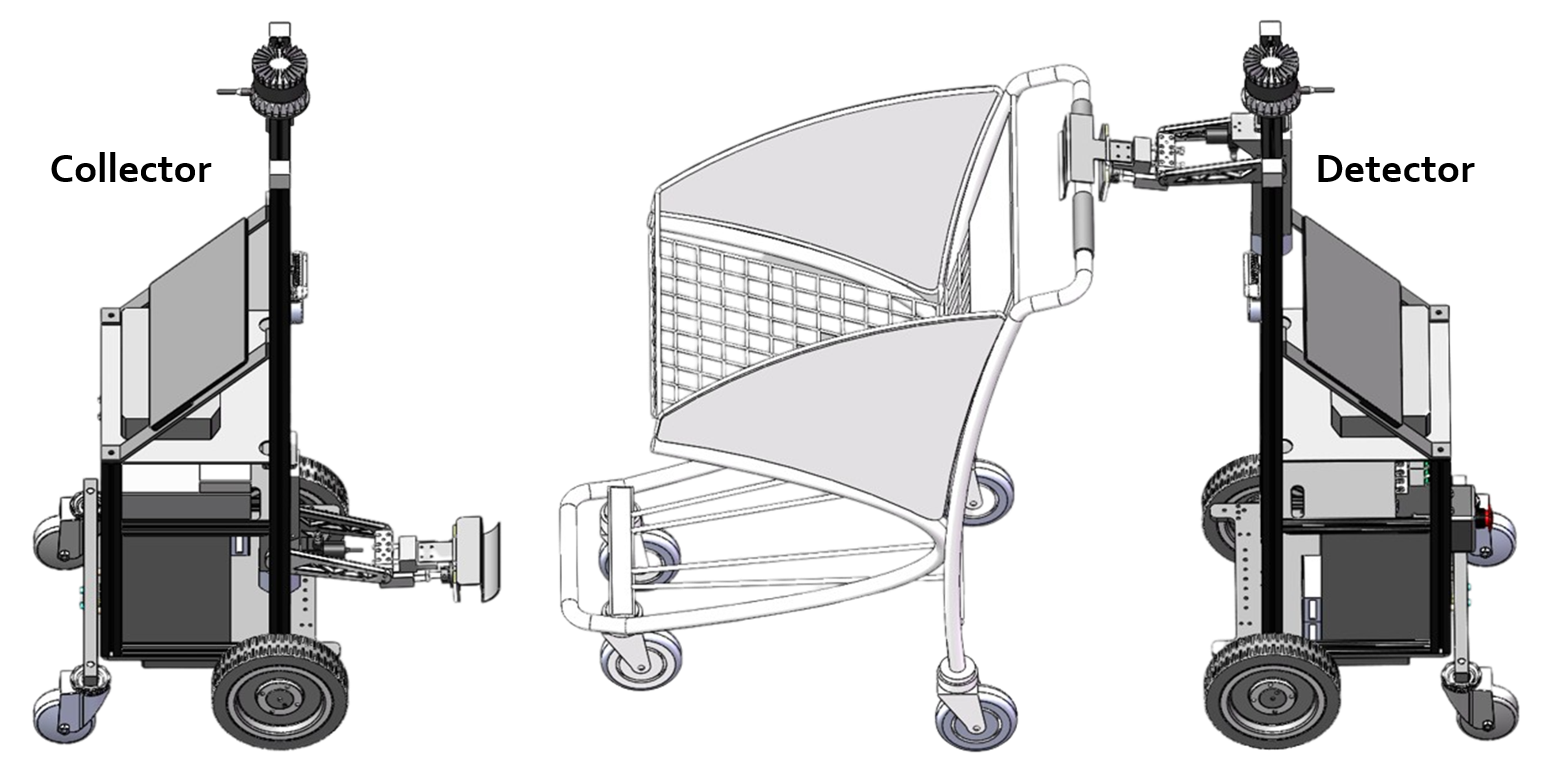}}
    \subfigure[]
    {\includegraphics[width=0.35\linewidth]{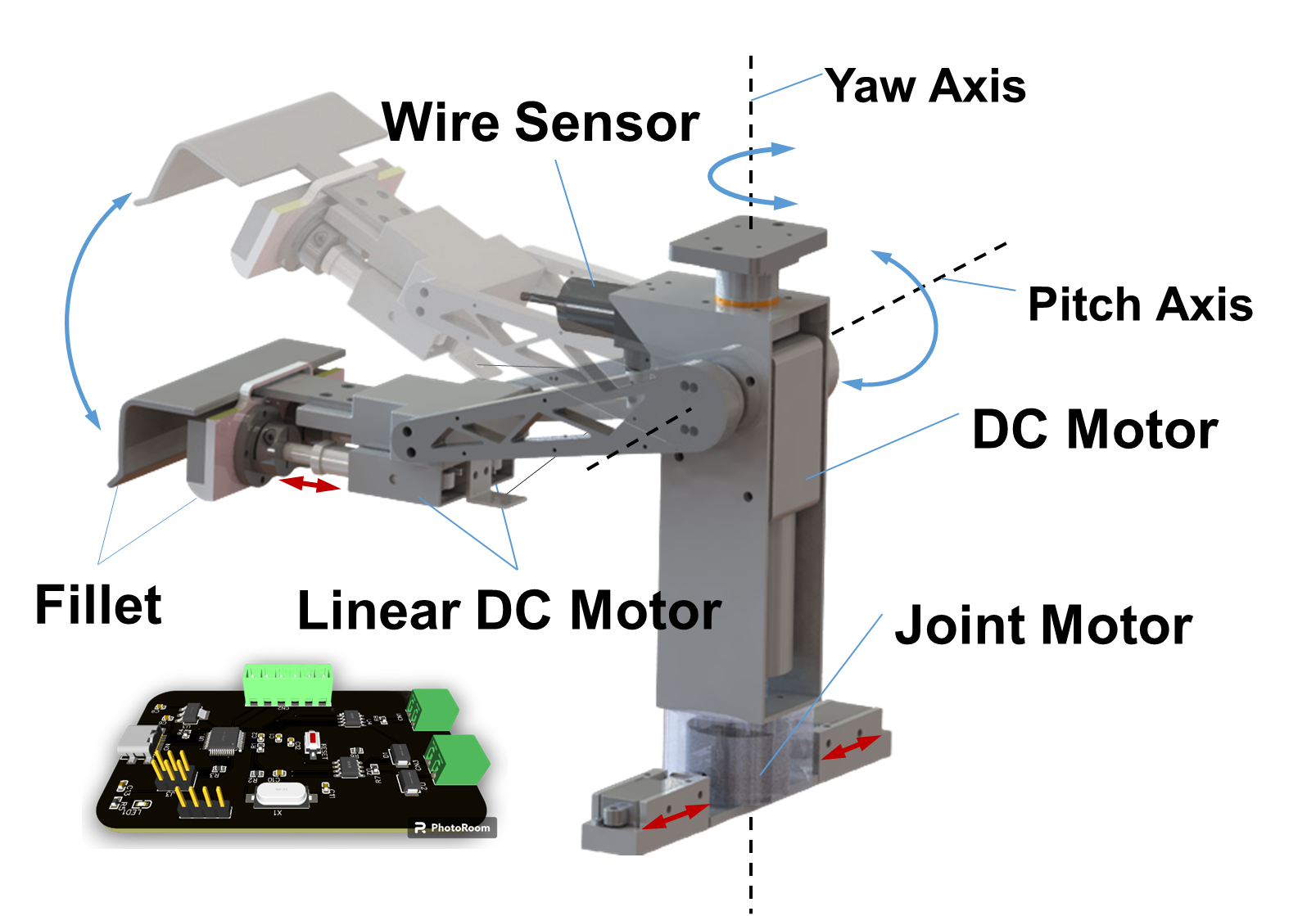}}
   
    \caption{
    The highly integrated hardware robotic system for trolley collection.
    (a) The Detector robot carrying a trolley to the Collector robot in the Docking stage.
    (b) The specially designed lightweight manipulator module to grab the beam of trolleys.
    }
    \label{fig:mech}
\end{figure*}

\begin{figure}[tb!]
    \subfigure[]
    {\includegraphics[width=0.48\linewidth]{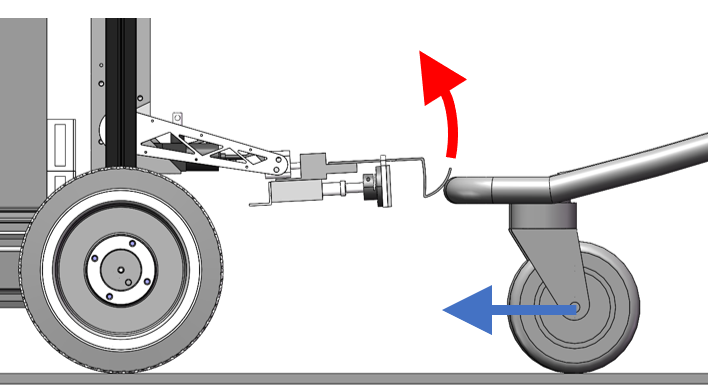}}
    \subfigure[]
    {\includegraphics[width=0.48\linewidth]{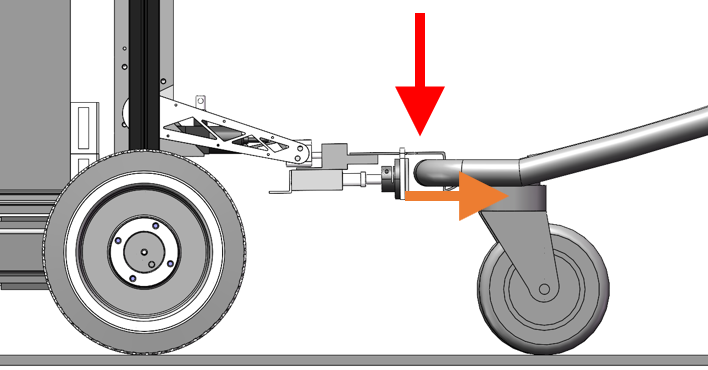}}
  \caption{The movement of the gripper when the detector robot grabbing the trolley approaches the collector robot.  }
  \label{fig:docking_mech}
\end{figure}

\subsection{Vision-based Approaching and Docking}
In the Approaching and Docking stages, the robot navigates to the target object localized from the onboard sensor observation.
Although the target objects are distinct in the two stages, the processes are conceptually similar to the tasks widely employed in robot charging\cite{mathew2015multirobot} and docking\cite{narvaez2020autonomous}.
The vision-based approach avoids the effect of global localization error that only utilizes the relative measurement to benefit the control accuracy.
The object detection and segmentation are the essential basic modules to determine the virtual target of the robot.
In practice, precise robot-object contact control\cite{he2023image} and outdoor wire inspection\cite{xing2023autonomous} rely on extracted edge feature from the image segmentation, while \cite{chen2021virtual} and \cite{pan2021} estimate object pose via point cloud registration.

\subsection{Optimization-based Control}
Recently, optimization-based strategies have gained their prevalence in robot planning and control and show promising performance in mobile manipulation\cite{pankert2020perceptive, peric2021direct}.
The CLF and CBF are introduced in optimization for safety-critical controller\cite{ames2016control, jankovic2018robust}.
It benefits realizing the optimal performance of robotic systems while ensuring the system stays in the safe set by inequalities from a CLF and a CBF.
There are also breakthroughs in tackling robot real-time safety-certificate control problems with CLF and CBF in optimization, such as robot formation control\cite{wang2017safety}, collision avoidance\cite{bena2023safety}, and legged-robot locomotion\cite{ames2019control}. 
We formulate the robot base navigation problem in the Approaching and Docking stage under the CLF-CBF-QP framework, to guarantee the system stability with the FoV constraint.

    

\section{System Description}
\label{sec:SD}
The proposed system is capable of efficiently catching the trolleys and sequentially stacking multiple luggage trolleys into a queue. 
To this end, an effective docking mechanism and an autonomy framework are designed. 

\subsection{Mechanical Design}
As shown in \figref{fig:mech}, the two mobile robots are with homogeneous structures. 
Specifically, their chassis uses two servomotors as the Driver wheel, each providing 32.75 $\text{N}\cdot \text{m}$ torque.
The Driven wheel is constructed by two passive omnidirectional wheels and utilizes a torsion beam suspension for shock absorption. 

As for the manipulator, existing designs either as a gripper to maintain a rigid body passively for linking it and the robot \cite{xiao2022robotic}, or as a passive joint to give the adjustments in the dynamic state for trolley transportation \cite{xia2023collaborative}. 
It just considers the specific, singular task.
And neither of them can fit in the whole pipeline and guarantee an accurate docking.
Therefore, it is necessary to design a state-switchable manipulator and a docking mechanism.
The details are as follows:
\begin{figure*}[tb!]
    \centering
    \includegraphics[width=0.98\linewidth]{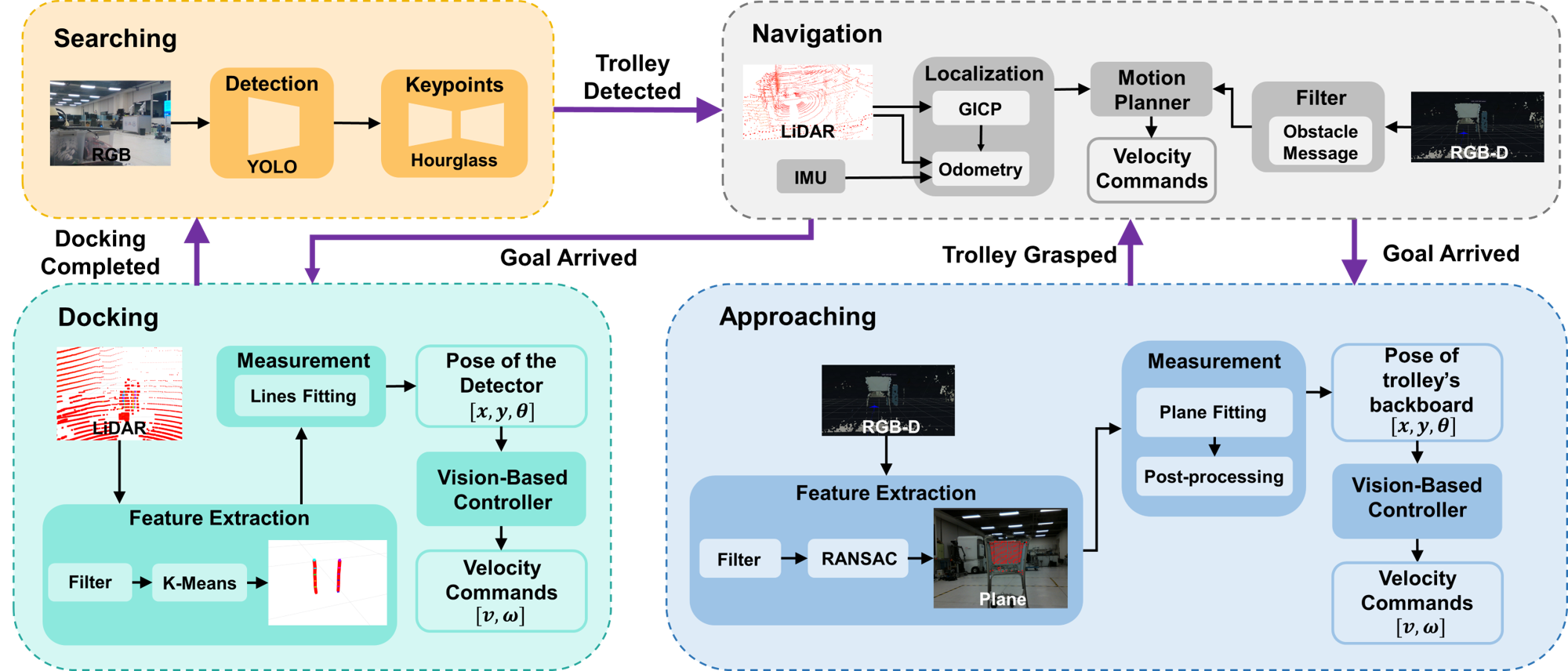}
    \caption{A diagram of our proposed autonomy framework.}
    \label{fig:framework}
\end{figure*}

\subsubsection{Manipulator}
The manipulator has a 2-finger end effector and 2 joints: a pitch axis and a yaw axis, as shown in \figref{fig:mech}(b). 
The pitch axis of the manipulator is driven by a DC motor and equipped with a wire sensor, allowing us to measure the manipulator's rotation angle around the pitch axis by the length of the wire.
The rotation about the yaw axis is controlled by the joint motor and undergoes two distinct state changes between the collection and transportation tasks. 
In the Docking stage, it is locked to maintain a rigid body, and in the Transportation stage, it is enabled to achieve a free state. 
To switch the manipulator's state, the joint motor responds to center the manipulator, aligning it with the latch, while the linear motors control the latch lock.
Finally, the end effector is driven by two linear motors that regulate its opening and closing. 
A fillet with a 5 cm radius is designed to enhance the fault tolerance when lowering the end effector to grab the beam of a trolley.
In addition, the control command is transmitted to the embedded board from the mini PC instructions directly via a serial port by utilizing Virtual Com.

\subsubsection{Docking Mechanism}
The Docking stage requires the Detector robot to arrange multiple luggage trolleys tightly into a queue through mobile manipulation in succession. 
We encounter two typical situations.
The first situation involves docking a trolley to an existing trolley queue led by the Collector robot.
When a trolley is already positioned in the queue, the existing trolley will serve as a guide for the succession (i.e., stackability) leading to the tolerance of about 8 cm lateral error and 10$^\circ$ orientation error.

In the other situation, there is no trolley in front as the guide, and meticulous collaboration between the two robots becomes necessary, as depicted in \figref{fig:mech}(a).
On the Collector robot's end-effector's tip, we designed a sleek, arc-shaped guiding structure. 
When the Detector robot propels the trolley toward the Collector robot and the trolley's front beam contacts this guide, the front beam applies a force parallel to the moving direction at the contact point.
The arc-shaped guiding structure can apply the reaction force perpendicular to the ground, prompting the manipulator to ascend.
As the trolley proceeds forward shown in \figref{fig:docking_mech}(b), the manipulator drops due to the gravity.
Simultaneously, the Detector robot assesses the pose to verify if the trolley is correctly aligned.
When the trolley reaches the correct alignment, the Collector robot drives the manipulator to catch the trolley.

\subsection{Autonomy Framework}
We design a hierarchical autonomy framework illustrated in \figref{fig:framework} dividing the TCP into four stages.
The system begins at the Searching stage and performs the collection task in a loop as the finite state machine until collecting the designated number of trolleys.
In the Searching stage, the Detector robot's RGB camera is activated to detect the trolley in the view. 
If an idle trolley is detected, the robot will measure its relative 6D pose by key points and set the pose as the goal of the Navigation stage.

In the Navigation stage, the Detector robot performs point-to-point navigation while ensuring collision avoidance.
As a transition stage, it is triggered according to the trolley status as the purple arrow indicating in \figref{fig:framework}, and works in the manner of a general navigation framework for the differential drive robot.
The LiDAR and IMU are utilized for global localization through the Simultaneous Localization and Mapping (SLAM) providing the occupancy map for planning.
In addition, the LiDAR and camera perceive the dynamic obstacles for safety.
Finally, the motion planner calculates the velocity commands to track the waypoints that are generated by the global planning on the established map.

In the Approaching stage, the Detector robot uses an RGB-D camera to perceive the relative pose of the trolleys in real-time. 
The backboard plane feature with a fixed shape is extracted and fitted to estimate the target pose for the tracking controller.
Then the vision-based controller calculates the velocity command and steers the Detector robot to minimize the pose difference in consideration of FoV. 
As the pose converges to the target, the Detector robot drives its manipulator to catch the trolley. 
After verification of the trolley status, the Detector robot inquires about the global pose of the Collector robot by socket UDP (User Datagram Protocol) as the navigation goal. 

The Docking stage is in the same manner as the Approaching stage, but the point cloud from LiDAR is utilized for relative measurement.
Special reflective markers on the Collector robot are regarded as the feature that can be filtered by their reflectivity.
Once the docking process is completed, The Detector robot will transition to search and collect the next trolley.

\section{Proposed Tracking Controller}
\label{sec:CTR}

The vision-based controller plays a crucial role in providing precise robot base control for trolley collection. 
Specifically, we develop a controller tailored for this constrained navigation problem using local robot observations.
We leverage the CLF-CBF-QP framework\cite{ames2019control} to compute in real-time and ensure accurate tracking while considering the FoV. 
In this section, we briefly introduce the fundamentals of CLF and CBF. 
Then, we define the control problem based on the kinematics model. 
Finally, we present the formulation of the QP-based controller.

\subsection{Notations}
To facilitate the understanding of formulas, common notations in this paper are listed here, such as vectors by bold-lowercase letters, e.g., $\mvc{x}$, matrices by bold-uppercase letters, e.g., $\mvc{A}$, and sets by calligraphic-uppercase letters, e.g., $\mathcal{S}$.
The transpose of a vector or a matrix is denoted as $\mvc{x}^\top$ or $\mvc{A}^\top$.
Plain letters with blankets mean the Cartesian frame such as the global frame $\mfra{O}$.
The superscript describes the coordinate frame of a variable, e.g., $\theta^R_T$ denotes angle $\theta_T$ w.r.t the frame $\mfra{R}$.

\subsection{Preliminary}
Combining CLF and CBF in optimal control effectively enforces the stability and safety of a nonlinear affine control system in the form of
\begin{equation}
    \dot{\mvc x} = f(\mvc x)+g(\mvc x)\mvc u
\end{equation}
where $\mvc x \in \mathcal{D} \subset \mathbb{R}^m $ represents the system state, $\mvc u \in \mathcal{U} \subset \mathbb{R}^n$ represents the control input, and the dynamical equation is locally Lipschitz.
Considering the control objective of asymptotically stabilizing the nonlinear system to the zero-point i.e., $\mvc x(t) \rightarrow 0$, a feedback control law that drives a CLF to zero guarantees the system stability.\\
\textbf{Remark 1 }
\textit{A control Lyapunov Function $V: \mathcal{D} \rightarrow \mathbb{R}_{\geq 0}$ is continuously differentiable and positive-definite. 
The system converges to zero-point with $V=0$ if there exists the control law satisfying:}
\begin{equation}
    \inf _{\mvc u \in \mathcal{U}} \dot{V}(\mvc x, \mvc u) \leq-\gamma(V(\mvc x))
\label{eq:clf}
\end{equation}
\textit{where $\mathcal{D}$ is the function definition domain, $\gamma: \mathbb{R}_{\geq 0} \rightarrow \mathbb{R}_{\geq 0}$ belongs to the class-$\mathcal{K}$ function. }



CBF is introduced to ensure the forward invariance of a safe set that any trajectory starting inside an invariant set will never reach the complement set.
The safe set $\mathcal{C}$ is defined on a continuously differentiable function $h: \mathcal{D} \subset \mathbb{R}^m \rightarrow \mathbb{R}$ i.e., the super-level set $\mathcal{C}= \{\mvc x \in \mathbb{R}^m \mid h(\mvc x) \geq 0\}$.
Then, $\mathcal{C}$ is forward invariant when $\mvc x(t) \in \mathcal{C}$ for all $t >0$ and any initial state $\mvc x(0) \in \mathcal{C}$.
Consider the aforementioned safe set $\mathcal{C}$ of function $h$, the function is called a Control Barrier Function if there exists an extended class-$\mathcal{K}_\infty$ function $\pi$ satisfying: 
\begin{equation}
    \sup _{\mvc u \in \mathcal{U}}  \dot{h}(\mvc x, \mvc u) \geq -\pi(h(\mvc x))
    \label{eq:cbf_set}
\end{equation}
\textbf{Remark 2 } 
\textit{The safe set $\mathcal{C}$ is forward invariant if there exist a CBF $h$ on $\mathcal{C}$ deriving any Lipschitz continuous controller $\mvc u$ that satisfies the following condition 
\begin{equation}
    \dot{h}(\mvc x, \mvc u)+\pi(h(\mvc x)) \geq 0, \forall \mvc x \in \mathcal{D}
\end{equation}}

The CLF and CBF can be incorporated together in optimization based controllers as inequality constraints of the control input.  
Based on the Quadratic Programming, such controllers calculate the current optimal control input of a safety-critical nonlinear system.
Now we formulate the general optimization problem of a QP-based controller as follows:
\begin{equation}
    \begin{aligned}
    \mvc u^* &= \underset{\mvc u, \delta}{\operatorname{argmin}} \  \frac{1}{2} \mvc u^\top \mvc H \mvc u + \delta^2 \\
    \text { s.t. } &  \dot{V}(\mvc x, \mvc u) + \gamma(V(\mvc x)) \leq \delta \\
    & \dot{h}(\mvc x, \mvc u)+\pi(h(\mvc x)) \geq 0 \\
    & \mvc u \in \mathcal{U}
    \end{aligned}
\label{eq:qp}
\end{equation}
where $\mvc H$ is a positive definite weight matrix, and $\delta$ is a variable that relaxes the CLF to ensure the QP has a solution.
When the function $\gamma$ and $\pi$ are designed as positive scalars for simplicity, the CLF constraint provides an exponential-decay upper bound which will converge to zero, and the CBF constraint provides an exponential-decay lower bound.

In this work, the system dynamics is based on the analysis of the rigid body motion in a 3-dimensional space. 
With calibration and setting the position of the sensors aligned with the robot rotation axis, the whole robot can be described as a rigid body with the same body-fixed frame.
Consider a rigid body moving with its body-fixed coordinate frame $\mfra{i}$, the pose of frame $\mfra{i}$ w.r.t the global coordinate frame is represented by the homogeneous transformation matrix $\mvc T^O_{i}  = (\mvc R^O_{i}, \mvc p^O_{i}) \in SE(3)$, where $\mvc R^O_{i} \in SO(3)$ denotes the rotation matrix and $\mvc p^O_{i} \in \mathbb{R}^3$ denotes the position.
Then, the rigid body motion represented by the body velocity $\mvc V_i \in se(3)$ is calculated by taking the time derivative: 
\begin{equation}
    \dot{\mvc T}^O_{i} = \mvc T^O_{i} [\mvc V_i]= \mvc T^O_{i}
    \begin{bmatrix}
    [\omega_{i}]_{\times} & v_{i} \\
    0 & 0
    \end{bmatrix}
\end{equation}
where $[\mvc V_i]$ is the body velocity in the matrix representation, and $[\cdot]_\times$ represents the skew-symmetric matrix of a vector.
When another moving rigid body with frame $\mfra j$ is introduced, the motion of frame $\mfra i$ can be described w.r.t frame $\mfra j$ called the relative motion.
According to the formulation $\mvc T^j_{i} = \mvc {T}_{O}^{j} \mvc T_{i}^{O}$ of changing the reference frame from coordinate frame $\mfra {O} $ to $\mfra {j}$, the relative motion of frame $\mfra i$ in $\mfra j$ is driven as following.
\begin{equation}
    \dot{\mvc T}^j_i = -[\mvc{V}_j] \mvc T^j_{i} + \mvc T^j_{i} [\mvc{V}_{i}]
\label{eq:rrm}
\end{equation}


\subsection{Problem Definition}


Considering the robot inertia frame $\mfra{R}$ is attached to its rotation center, the model of the differential drive wheeled mobile robot is formulated as follows:
\begin{equation}
\label{eq:kinematic}
    \begin{cases}
        \dot{x} = v\cos{\theta} \\
        \dot{y} = v\sin{\theta} \\
        \dot{\theta} = \omega
    \end{cases}
\end{equation}
where $ [x , y] \in \mathbb{R}^2$ denotes the Cartesian coordinates on plane and $\theta \in \mathbb{R}$ denotes the heading angle.
The robot model is analyzed as a velocity-controlled unicycle model with the control input $\mathbf{u}=[v, \omega]$ consisting of two variables, linear velocity, and angular velocity respectively.
In addition, due to the control input limitation by the chassis and engine for real robots, the velocity is continuous and bounded by $|v| \leq v_{max}$ and $|\omega| \leq \omega_{max}$, where $v_{max}$ and $\omega_{max}$ are the velocity upper bounds.

\begin{figure}[t!]
    \centering
    \includegraphics[width=0.75\linewidth]{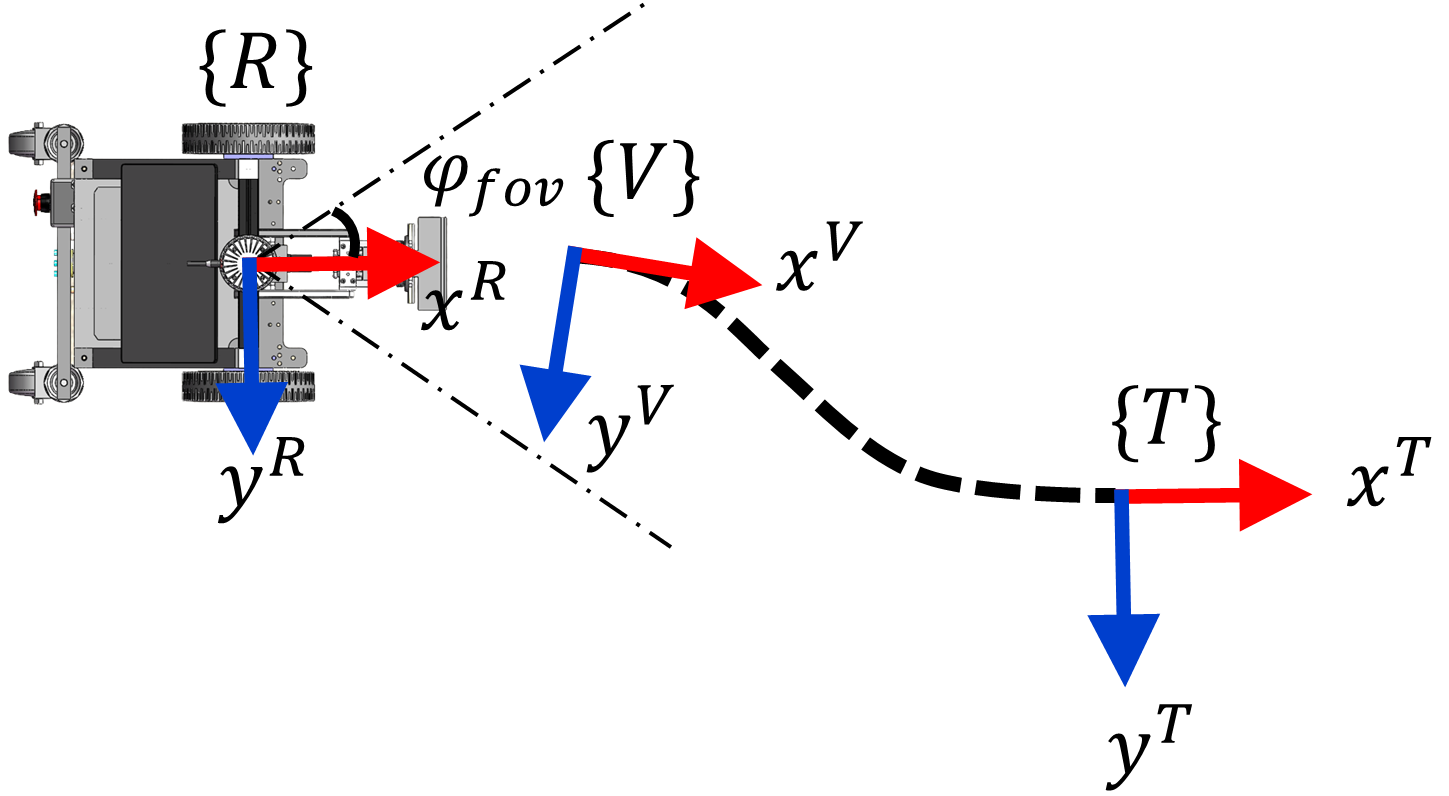}
    \caption{The illustration of robot target object approaching under relative measurement. The blue block is the robot, the yellow triangle is the target and the green frame is a virtual target moving along the planned green path.}
    \label{fig:ctr_model}
\end{figure}

The control objective is to navigate the robot to the target object in both the Approaching and Docking stages.
As shown in \figref{fig:ctr_model}, the relative pose of the target object frame $\mfra T$ w.r.t the robot frame $\mfra R$, which is measured through sensors as $\mvc x^R_T = [x^R_T, y^R_T, \theta^R_T]$ in the 2-dimensional plane.
To evaluate the performance, the steady-state error consisting of the normal error $e_y$, the lateral error $e_y$, and the orientation error $e_{\theta}$ is the relative pose $\mvc x^R_T(t_p)$ measured at the timestep $t_p$.

Assuming that a known trajectory plans the robot's base motion in the free space without obstacles, we define the following problem and propose a control method for it.\\
\textbf{Problem 1} 
\textit{
According to the robot's local observation $\mvc x^R_T$, solve the control input $\mvc u$ for the robot to move along the known trajectory $s(t)$ such that the steady-state errors converge to zero asymptotically.}


Considering the aforementioned trajectory tracking problem, the planned trajectory can be viewed as the relative pose $\mvc x^T_V$, a virtual target with body-fixed frame $\mfra V$, and the nominal velocity of $[v_V, \omega_V]$.
Then according to (\ref{eq:rrm}), the robot $\mfra R$ tracks the trajectory by minimizing the error, the relative configuration $\mvc x = \mvc{x}^R_V = [x^R_V, y^R_V, \theta^R_V]^\top$, which is formulated as following dynamics:
\begin{align}
\label{eq:err_dy}
    \dot{\mvc x} =
    \begin{bmatrix}
        y_V^R \omega_R  + v_V \cos{\theta_V^R} - v_R\\
        -x_V^R \omega_R +  v_V \sin{\theta_V^R} \\
        \omega_V - \omega_R
    \end{bmatrix}
\end{align}
where the $\mathbf{u}=[v_R, \omega_R]$ is the control input.

For planning the robot state to the goal, a twice differentiable trajectory of $s(t):[0,t_p]\rightarrow \mathbb R^3$ is valid in consideration of the Boundary Value Problem.
We use quintic polynomials represented by:
\begin{equation}
 \left\{
\begin{array}{lrc}
    x(t) = a_0+a_1t+a_2t^2+a_3t^3+a_4t^4+a_5t^5 \\
    y(t) = b_0+b_1t+b_2t^2+b_3t^3+b_4t^4+b_5t^5
\end{array}
\right.
\end{equation}
which further defined the velocity and the acceleration through the first derivative and the second derivative of time.
The coefficients $a_k$ and $b_k$ are solved by the boundary condition equations, the boundary positions are $s(0)$ and $s(t_p)$, the boundary velocities are $\dot{s}(0)$ and $\dot{s}(t_p)$, and the boundary accelerations are $\ddot{s}(0)$ and $\ddot{s}(t_p)$.


\subsection{Optimization Based Control}

The proposed tracking controller is based on the CLF-CBF-QP in (\ref{eq:qp}) to ensure the FoV constraint and the convergence to the planned trajectory.
First, to minimize the difference between the actual and the planner states, we derive the CLF candidate with the relative configuration relationship $\mvc x$ as follows:
\begin{equation}
    V(\mvc{x}, \mvc{u})=\frac{1}{4}(\mvc x^\top \mvc H \mvc x)^2
\end{equation}
where $\mvc H$ denotes the positive diagonal weight matrix.
The derivative of $V$ w.r.t time along the robot's trajectory is 
\begin{equation}
    \begin{aligned}
    \dot{V}(\mvc x, \mvc u)= & (\mvc {x}^\top \mvc H  \mvc {x} ) \mvc {x}^{\top} \mvc H \dot{\mvc{x}} \\
    \end{aligned}
\end{equation}
where we introduce the error dynamics (\ref{eq:err_dy}).
Choosing the function $\gamma(V)$ in (\ref{eq:clf}) a positive scalar, the CLF constraint in QP is then derived as:
\begin{equation}
    \dot{V}+ \mu V \leq \delta
\end{equation}
where $\mu>0$ is a scalar parameter and $\delta>0$ is a slack variable for solvability.


Secondly, we use CBF in optimization to guarantee the robot does not lose sight of the target object as shown in \figref{fig:ctr_model}.
The angle between the relative position $\mvc p^R_T$ and the robot orientation in robot frame $\mfra R$ is bounded in an invariable interval:
\begin{equation}
\label{eq:bra}
    |\varphi| = |\arctan(\mvc p^R_T)| < |\varphi_{fov}|
\end{equation}
where function $\arctan(\mvc p) = y/x$ denotes relative angle and the FoV angle $\varphi_{fov}>0$ is determined by sensors. 
$\varphi$ is calculated from the robot local observation.
Then, we describe the angle constraints as the CBF candidate:
\begin{equation}
    h(\varphi) = \frac{1}{2} (\varphi_{fov}^2-\varphi^2 )
\end{equation}
where $\{\varphi \mid h\left(\varphi\right) \geq 0\}$ constructing a safe set according to (\ref{eq:cbf_set}). 
In the same manner, choosing the function $\pi(h)$ a scalar, we derive the CBF constraints:
\begin{equation}
    \dot{h} + \lambda h \geq 0 .
\end{equation}
where $\lambda$ is parameter.
Furthermore, we formulate the CBF constraint by differentiating $h(\varphi)$ w.r.t. time:
\begin{equation}
    \dot{h}(\varphi,\mvc u) = -\varphi \dot{\varphi} = \frac{\varphi}{\rho^2} [{(y_T^R)^2 \omega_R - y_T^R v_R} + (x_T^R)^2 \omega_R]
\end{equation}
where $x^R_T$ and $y^R_T$ are from the presence robot local observation.

Finally, we introduce the inequality form of CLF and CBF constraints into the QP to solve the vision-based tracking control problem as follows:
\begin{subequations}
    \begin{align}
        \min _\mvc{u} & \quad \mvc{u}^\top \mvc Q_u \mvc u + c_\delta \delta^2 \label{eq:opti_cost}\\
        \text { s.t. } 
        &\mvc{u} \in \mathcal{U} \label{eq:u_feas}\\
        &\dot h(\varphi, \mvc{u} )+\lambda h(\varphi) \geq 0 \label{eq:dclf}\\
        & \dot V(\mvc x, \mvc u)+\mu V(\mvc x) \leq \delta \label{eq:dcbf}
    \end{align}
\end{subequations}
where the positive definite matrices $\mvc Q_u$ are coefficients matrices measuring the control costs, and parameter $c_\delta$ is the weight to penalize the slack variable in CLF. 
In addition, the constraint \eqref{eq:u_feas} represents the admissible control constraints, where the robots' velocities are bounded in the sense of $|v| \le v_{\textrm{max}}$, and $|\omega| \le \omega_{\textrm{max}}$, and accelerations are bounded via discretization as well $|{v} - v_0| \le a_{\text{max}}$, $ |\omega -\omega_0| \le \alpha_{\text{max}}$. 
The QP problem is formulated by CasADi \cite{andersson2019casadi} and solved with IPOPT \cite{biegler2009large} on the robot's onboard computer in real-time. 
\section{Target Object Localization}
\label{sec:VT}

\begin{figure}[tp]
   \subfigure[]
    {\includegraphics[width=0.48\linewidth]{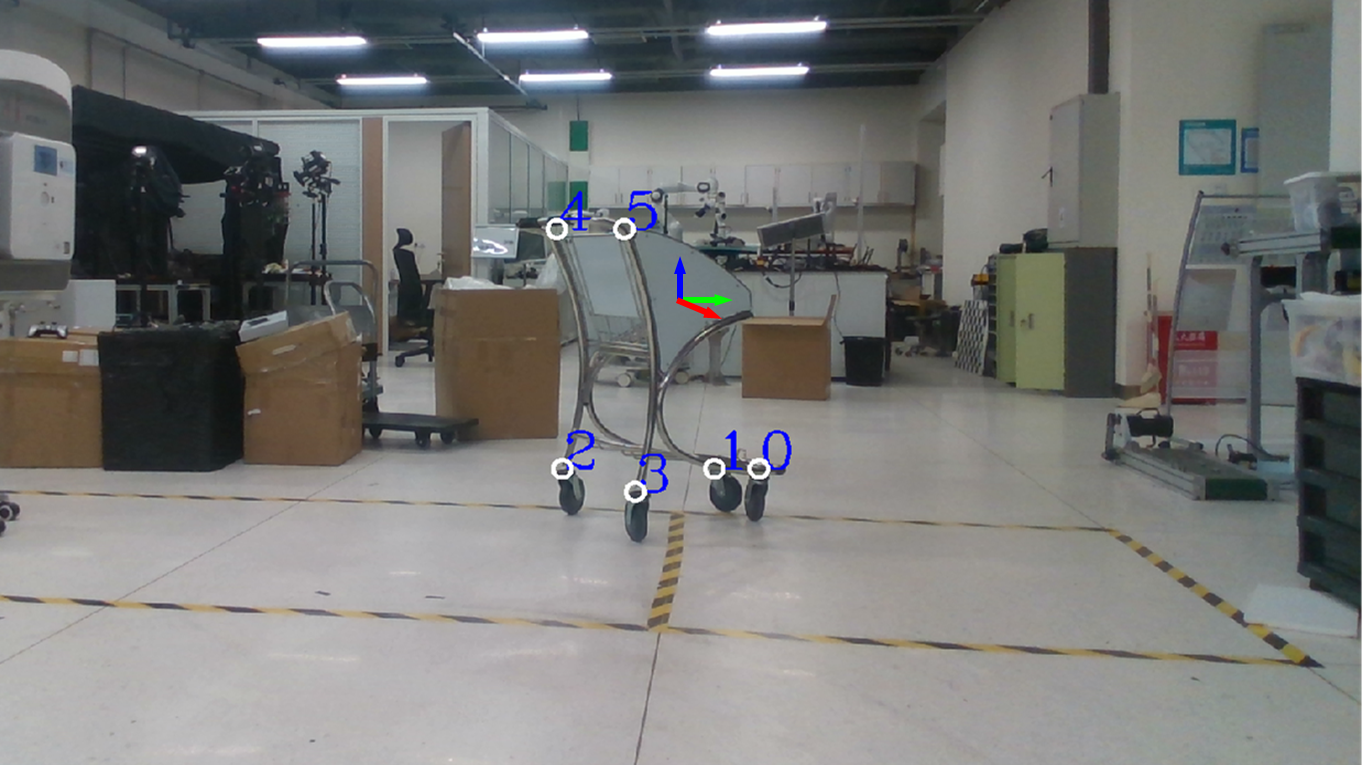}}
    \subfigure[]
    {\includegraphics[width=0.48\linewidth]{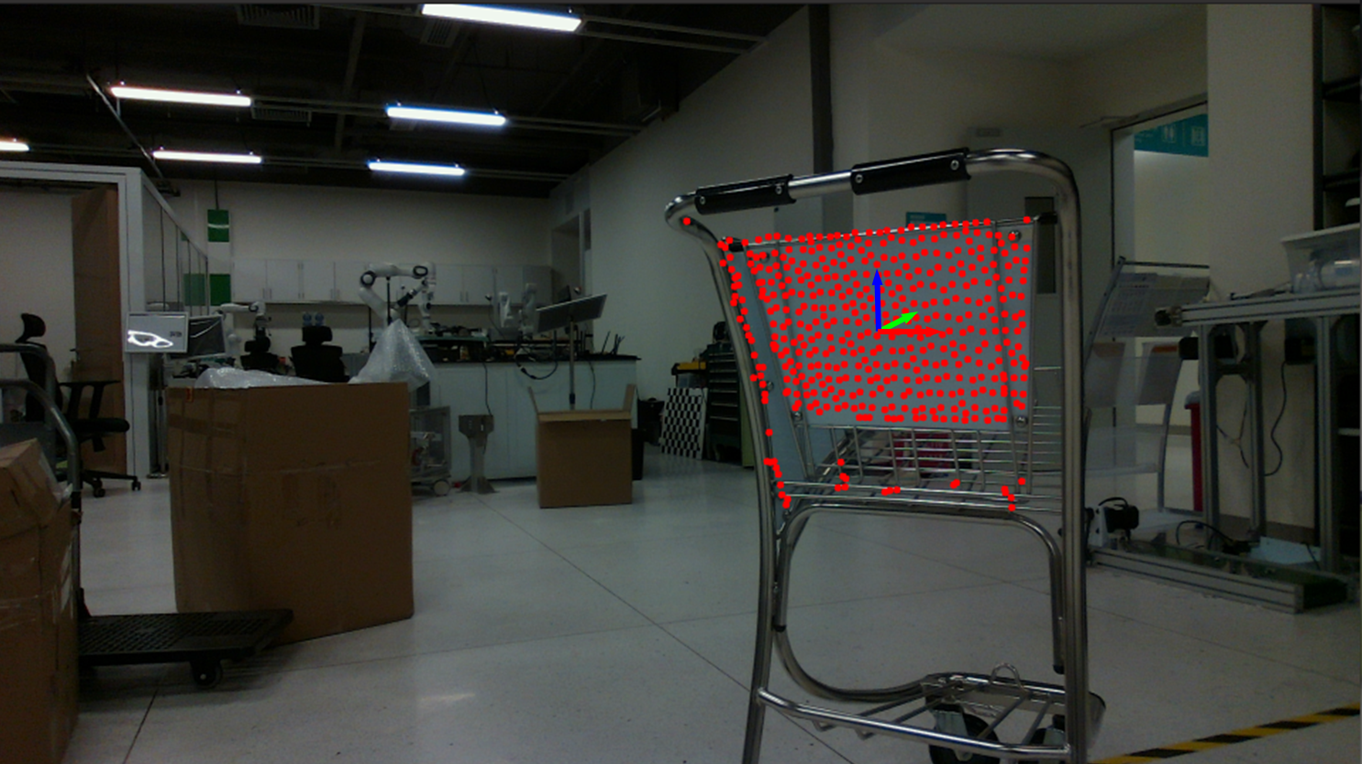}}
    
    \caption{
    The illustration of the course-to-fine pose estimation results. (a) At a long distance, the trolley is detected on an RGB image and localized by the keypoints features. (b) Precise localization through point cloud segmentation at a close distance.
    }
    \label{fig:pose_estimate}
\end{figure}
Different onboard sensors are incorporated for visual perception of the distinct object features providing the goal pose for planning and control.
Based on our previous work\cite{xiao2022robotic}, learning-based detection and feature corresponding pose estimation methods are employed in different stages.

\subsection{Trolley Localization}
We implement a course-to-fine perception approach on an RGB-D camera to detect and localize an idle trolley in sight. 
In the Searching stage, the Detector robot detects the idle trolleys and estimates their approximate pose with the RGB image, providing the coarse goal pose for robot navigation.
As the robot navigates to a closer distance of the target trolley, enough features are in sight.
Then, the depth point cloud is used for more accurate pose estimation in the Approaching stage. 

\subsubsection{Coarse Perception}
First, the YOLOV5 \cite{yolov5} network is fine-tuned for real-time trolley detection cropping the input image to the bounding box of a trolley.
Then, as shown in \figref{fig:pose_estimate}(a), the stacked Hourglass network model\cite{hourglass} predicts the 6 keypoints of the trolley.
Denoting the homogeneous coordination of each detected keypoint $i=0,1,...,5$ as $\mvc h^c_i$ in the image frame and $\mvc h^w_i$ in the real-world object frame, the camera perspective projection model describes the correspondence between them:
\begin{equation}
    s_i \mvc h^c_i = \mvc K 
    [\mvc R \ \mvc p]
    \mvc{h}^w_i , 
\end{equation}
where $s_i$ is the scale factor, $\mvc R$ and $\mvc p$ are the rotation matrix and the translation of the robot respectively, and $\mvc K$ is the intrinsic camera parameter matrix.
Finally, the relative pose between the robot and the trolley is estimated by solving the perspective-n-point (PnP) problem. 
According to the estimated relative pose in the camera frame, the navigation goal can be obtained by the localization information of the robot base.

\subsubsection{Fine Perception}
The plane segmentation and fitting are adopted to the point cloud from the RGB-D camera for the precise pose estimation of the trolley as shown in \figref{fig:pose_estimate}(b).
As the pre-processing, the original point cloud is filtered with a pass-through distance filter that remains the point cloud set of $\mathcal{P}_{f}=\{ \mvc p_f \mid -0.3\text{m} < y_{f} < 0.6\text{m},\ z_{f} < 2.5\text{m} \}$, where $\mvc p_f$ is the coordination of valid point in the camera frame.
Considering the plane feature of a trolley's backboard, we employ the Random Sample Consensus (RANSAC) algorithm from the PCL library to fit the parameters $A, B, C,$ and $D$ in equation $Ax_{f}+By_{f}+Cz_{f}+D=0$. 
Then, projecting to the X-Y plane w.r.t the robot frame $\{R\}$, the trolley's pose is represented by the centroid coordination of the backboard $\hat{\mvc{d}} = [\hat{x}^R_T, \hat{y}^R_T]^{\top}$ and the normal angle $\hat{\theta}^R_T = -\arctan(\frac{A}{C})$. 
     


When partially observing the trolley’s backboard, the RANSAC algorithm tends to incompletely segment the backboard causing a noticeable deviation between the estimated centroid coordination $\hat{\mvc d}$ and the actual centroid coordination.
To compensate for the FoV occlusion, we introduced a post-process that leverages the prior knowledge of the trolley's backboard geometry.
With the observed relative distance between the robot and the centroid $\hat{\rho}=\sqrt{(\hat{x}^R_T)^2+(\hat{y}^R_T)^2}$, we can estimate the length of the trolley backboard in sight $\hat{l}$ by the sine law:
\begin{equation}
    \frac{\hat{\rho}}{-\cos{(\hat{\theta}^R_T + 2\varphi_-\varphi_{fov})}}=\frac{\hat{l}}{2\sin{(\varphi_{fov}-\varphi)}}
\end{equation}
where $\varphi$ is the bearing angle between the robot and the trolley as in \eqref{eq:bra}, and $\varphi_{fov}$ is the FoV angle.
Finally, we can modify the observation $\hat{\mvc{d}}_{cp}$ with the view compensation as follows:
\begin{equation}
    \hat{\mvc{d}}_{cp} = \hat{\mvc{d}} + \frac{1}{2} (\hat{l}-l)[\cos{\hat{\theta}^R_T} ,\ -\sin{\hat{\theta}^R_T}]^{\top}
\end{equation}
where the length of the trolley's backboard $l$ is a constant.

\subsection{Robot localization}
In the Docking stage, the monocular camera is fully occluded by the caught trolley.
Two special reflective strips on the Collector robot are utilized as markers for docking localization.
The markers generate high-intensity point clouds in the 3D LiDAR scan that can be detected through a high-pass filter to segregate points exhibiting an intensity surpassing 4500 lux. 
The detected raw point cloud set is denoted as $\mathcal{P}_{sc} = \{(\mvc p_{sc}, I_{sc}) \mid I_{sc} > 4500 \ \text{lux} \}$, where $I_{sc}$ represents the reflectivity of each scanned point and $\mvc p_{sc}$ represents the Cartesian coordinates w.r.t the robot. 
The relative translation between the two robots is calculated as the mean vector of the $\mathcal{P}_{sc}$: 
\begin{equation}
   [\hat{x},\hat{y},\hat{z}]^{\top} =  \frac{1}{n}\sum^{n}_{i=0}{\mvc{p}_{sc,i}}
\end{equation}
Furthermore, by the KMeans clustering, the scanned points corresponding to the two markers are identified as two subsets of the original point clouds denoted as $\mathcal{P}_l$ and $\mathcal{P}_r$ respectively.
We calculate the mean vectors $\mvc{\bar{p}}_l$ and $\mvc{\bar{p}}_r$ of each subsets, and the norm vector $\mvc{v}$ indicating the orientation of the Collector robot is obtained by geometrical relationship:
\begin{equation}
    (\mvc{\bar{p}}_l-\mvc{\bar{p}}_r)\mvc{v} = 0
\end{equation}
where $\mvc{v}=[x_v,y_v,z_v]\in \mathbb{R}^3$.
Then, the bearing angle of the Collector robot w.r.t the robot is obtained by projecting the normal vector to the 2D plane. 
Finally, the docking goal pose is obtained in consideration of the queue length, where each collected trolley in the queue adds the offset distance $\xi$ alone the Collector robot's orientation.
\section{Experimental Results}
\label{sec:ER}
  
\subsection{Implementation Details}

\begin{figure}[tp!]
\centering
\includegraphics[width=1.0\linewidth]{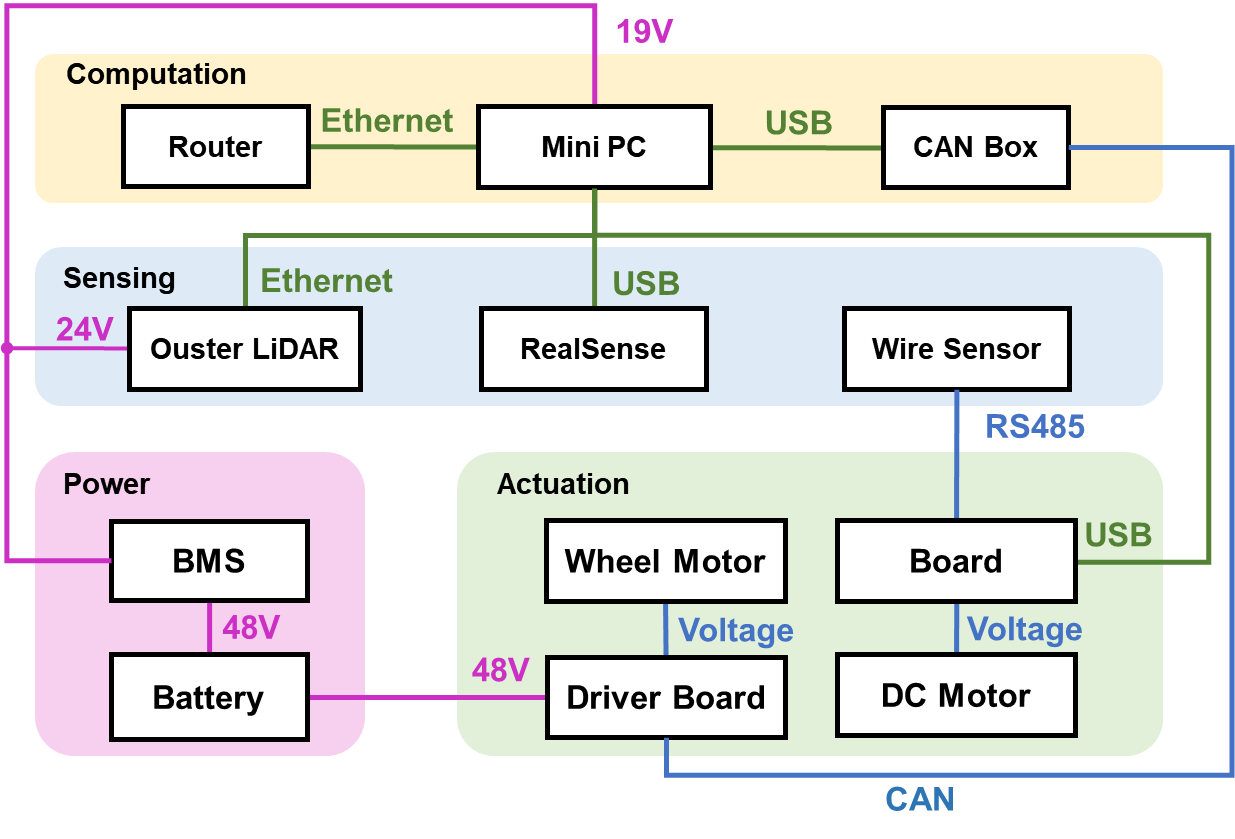}
\centering
\caption{The structural overview of robot electronics.}
\label{fig:electial_diagram}
\end{figure}

The electronic components of the two robot prototypes are identical as shown in \figref{fig:electial_diagram}. 
There are four main functional modules: the Power module, the Actuation module, the Sensors, and the Computation module.

\subsubsection{Power module}
It employs a 48V lithium iron phosphate battery, supplied to the chassis motors and complemented by an efficient battery management system (BMS) to supply 19V to the MiniPC and 24V to the LiDAR, respectively.
 
\subsubsection{Actuation module}
It employs 48V brushless DC servo motors to drive the chassis and DC motors to drive the manipulator.
With the CAN Box, the MiniPC can send commands to the driver board to control the movement of the chassis via USB.

\subsubsection{Sensors}
The 32-line Ouster LiDAR OS0 with the average deviation of 1 mm and the inner integrated 6-axis IMU are used for the robot's localization. 
The Realsense L515 camera with a vertical and horizontal angular coverage of $55^{\circ}$ and $70^{\circ}$ is used for object detection and obstacle avoidance.

\subsubsection{Computation module}
MiniPCs on the Detector and Collector are with distinct specifications in consideration of different functions. 
The Collector is equipped with a MiniPC with an i7-11657 CPU@4.70GHz, while the Detector uses a MiniPC with an i7-12900F CPU@5.10GHz and an RTX3060.
The autonomy framework is integrated through the Robot Operating System (ROS2 Galactic) with algorithms implemented in Python and C++.

\subsection{Robot Navigation}
We employ a similar framework as in \cite{xiao2022robotic} for safe robot navigation in a dynamic environment.
The implementation details are as follows:

\subsubsection{Localization}
We employ the FAST-LIO2\cite{fastlio} algorithm for SLAM based on the Ouster LiDAR. 
The initial pose of the odometry is solved by the Generalized Iterative Closest Point (GICP)\cite{gicp} algorithm that matches the LiDAR point cloud and the established 3D map point cloud.
For robustness, it relocalizes the robot to eliminate errors when the robot's movement surpasses predefined pose thresholds.

\subsubsection{Global planner}
We employ the A* algorithm on the 2D cost map to solve an optimal global path, a series of waypoints, due to the computation efficiency considering only indoor experiments.
The Octomap\cite{octomap} algorithm is utilized to diminish the height of a 3D map to 2D, and the corresponding cost map is yielded by expanding the obstacle pixels
Besides, the input point cloud data is filtered to a height between -1.0 m and 1.0 m to curtail the perturbations induced by the floor and ceiling.

\subsubsection{Obstacle Avoidance}
The nonlinear model predictive control method is implemented as the local planner for obstacle avoidance. 
The 2D distance to the obstacle, unknown point clouds on the map, is measured and formulated as constraints.
The point cloud from both the LiDAR and the RGB-D camera are collectively used with calibration, while only LiDAR data are utilized when the camera is blocked by the caught trolley.

\subsection{Trolley Pose Estimation Experiments}
\begin{figure}[t!]
    \centering
    {\includegraphics[width=0.98\linewidth]{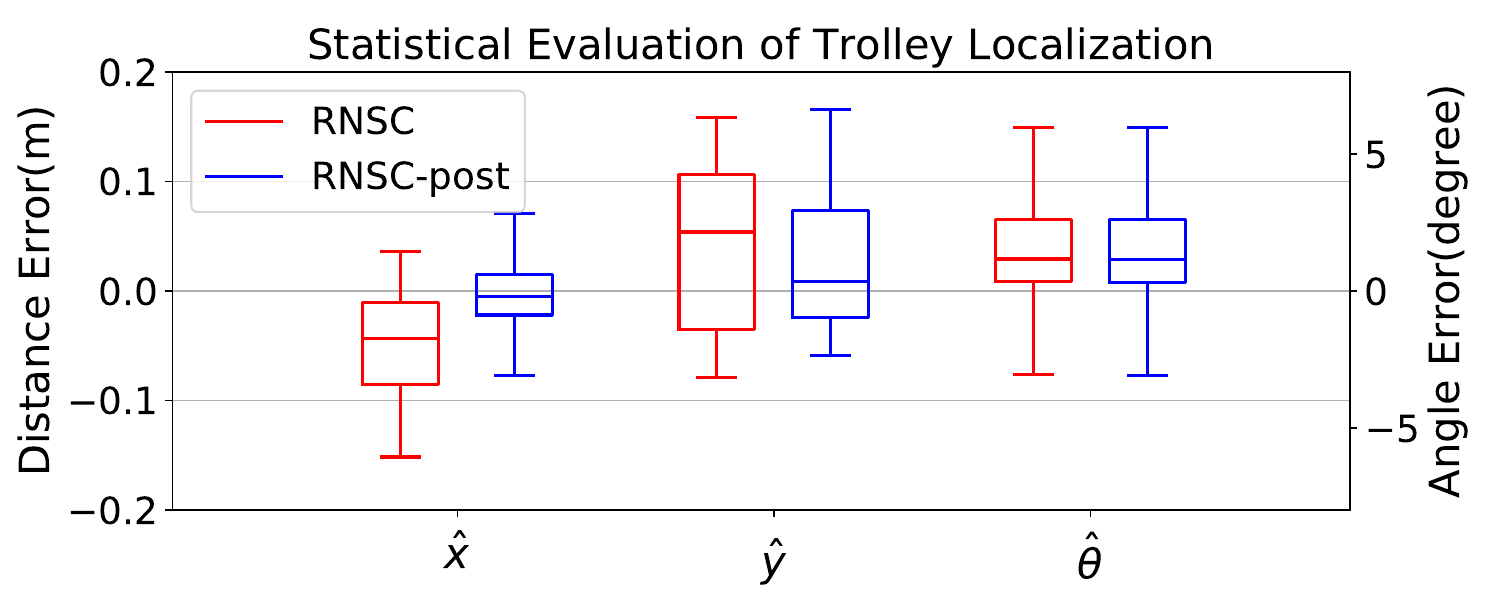}}\\
    \caption{
    The boxplot of the pose estimation error in real-world experiments. The RANSAC point cloud segmentation methods with or without post-process are compared in the 3 dimensions of trolley pose estimation. The line across a box shows the median.
    }
    \label{fig:boxplot}
\end{figure}
We conduct validation experiments of the proposed trolley localization method, and the ablation study comparing the localization performance with or without post-processing.
We set the Detector robot at 50 random poses relative to a trolley to measure the trolley's pose.
The OptiTrack Motion Capture system provided ground truth measurements, against which we assessed the average localization errors.
Three types of observation errors are analyzed: the longitudinal distance error $\hat{x}$, the lateral distance error $\hat{y}$, and the impact angle error $\hat{\theta}$.
These errors are computed as the difference between the observed value and the ground truth.

The statistical analysis of the observation errors is presented in \figref{fig:boxplot} as a boxplot. 
As a result, the median values of all these errors fall within the range of $\pm \text{0.1 m}$ or $\pm \text{5}^\circ$, attesting to the precision of our trolley localization method based on the RANSAC algorithm. 
This level of accuracy substantiates our method of segmenting the backboard of the trolley as a distinctive feature for accurate localization.
A notable enhancement is observed with the integration of the post-processing step. 
This refinement brings the median values of $\hat{x}$ and $\hat{y}$ closer to zero than the original method and significantly narrows their interquartile ranges, indicating enhanced precision through compensation for view constraints.
Meanwhile, the angle error $\hat \theta$, estimated by the normal vector of the trolley's plane, remains consistent, as it is not influenced by the compensation.

\subsection{Control Methods Evaluation}
\begin{table}[t!]
\begin{flushleft}
\setlength{\tabcolsep}{4.5pt}

\caption{Comparison of different control strategies over the 30 trolley docking experiments.}
\label{tab:compare}
    \begin{tabular}{lccccccc}
        \toprule
        & \multirow{2}*{Success} & \multicolumn{2}{c}{$e_{x}$(mm)} & \multicolumn{2}{c}{$e_{y}$(mm)} & \multicolumn{2}{c}{$e_{\theta}$(degree)}\\
        \cmidrule{3-8}
          &       & av.  & std  & av. & std & av. & std\\
        \midrule
        Nonlinear\cite{chen2021virtual} & 15          &  20.54         & \textbf{0.77} & \textbf{1.37} & 18.62           & 3.02          & 18.83\\
        MPC\cite{xiao2022robotic}   & 21                  & 5.09          & 7.56     & 16.32         & 38.07       & \textbf{1.88} & 3.11\\
        Ours                            & \textbf{30}  & \textbf{2.08} & 0.96    & 9.33          & \textbf{10.51}      & 2.12          & \textbf{1.55}\\
        \bottomrule
    \end{tabular}
    
\end{flushleft}
\end{table}

\begin{figure}[t!]
    \begin{minipage}[c]{0.64\linewidth}
    \subfigure[]{
        \includegraphics[width=0.98\linewidth]{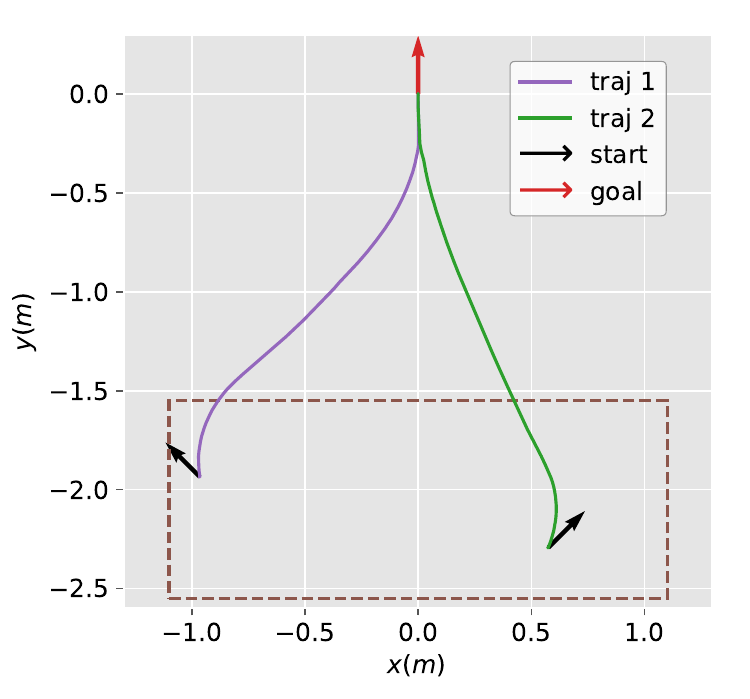}}
    \end{minipage}
    \begin{minipage}[c]{0.34\linewidth}
        \centering
        \subfigure[]{
        \includegraphics[width=0.96\linewidth]{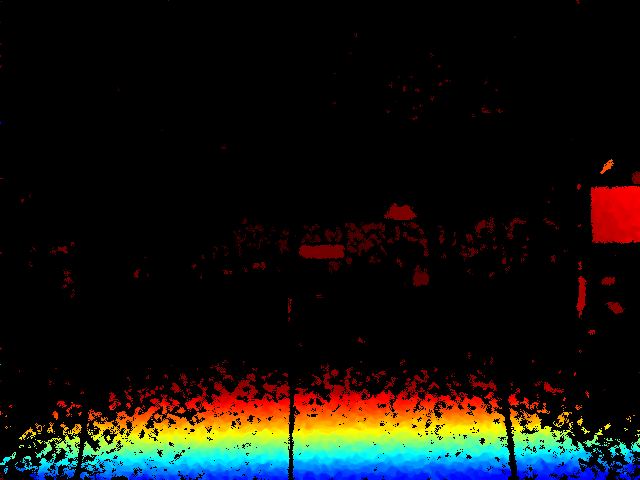}}
        
        \subfigure[]{
        \includegraphics[width=0.96\linewidth]{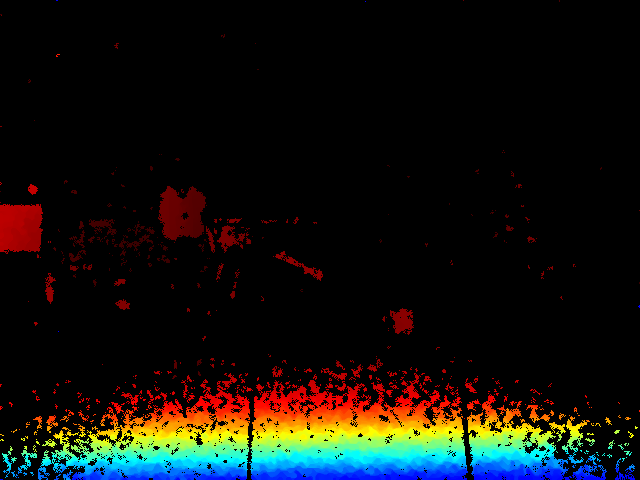}}
    \end{minipage}
  \caption{Two challenging situations in the control evaluation experiments. 
  (a) The curves record the actual trajectories of the robot steered by the proposed control method. 
  The black arrows are the robot's initial poses, and the red arrow is the target pose. 
  (b) The onboard RGB-D camera view at the beginning of Trajectory 1, where the bright pixels on the right side indicate the depth date of the trolley backboard.
  (c) The RGB-D camera view at the beginning of Trajectory 2.}
  \label{fig:first_exp_suitation}
\end{figure}

To assess the effectiveness and reliability of the proposed control method, comparison experiments are conducted on the robot prototype. 
In the experiments, the Detector robot undertakes the Approaching stage aiming to collect a static target trolley starting from 30 random initial poses.
Our method is benchmarked against two established control strategies: the Nonlinear method \cite{chen2021virtual} and the MPC method \cite{xiao2022robotic}. 
The experimental setup is depicted in \figref{fig:first_exp_suitation}(a) w.r.t the trolley's body-fixed frame.
The relative position of the robot is confined in the region (dashed box) of $\text{-1.0 m} \leq x \leq \text{1.0 m}$, $\text{-2.5 m} \leq y \leq \text{-1.5 m}$ ensuring the trolley is within sight at the start. 
 the same trolley localization algorithm is implemented across all methods to eliminate localization perturbation as a variable, and the results are evaluated against the ground truth.

The comparison results are presented in Table \ref{tab:compare} including success times and the terminal control errors in the successful experiments.
Considering the physical limitation of the manipulator and the trolley, Success is quantified by the robot's ability to reach a specified relative position within a tolerance of $\rho< \text{30 mm}$ for position and $\lvert \theta \lvert < \text{5}^{\circ}$ for orientation, enabling the manipulator to grasp the trolley.
In the successful experiments, the relative pose between the robot and the trolley at the terminal state is defined as control errors, which are compared by the mean and standard deviation values of the 3 groups.
The magnitude of $e_x$ and $e_\theta$ directly affect the manipulator to catch a trolley and the magnitude of $e_y$ influences the success of the Docking stage because the contact point is fixed and the error will impact the control in the Docking stage.

As a result, our method completes all 30 experiments, outperforming the other methods, especially in challenging scenarios.
The MPC and Nonlinear methods typically fail when requiring the control near the limit of state and input bounds, particularly when there is a significant difference between the initial pose and the goal.
While each approach demonstrated small enough angular errors $e_\theta$, our method significantly reduces $e_x$ and achieves competitive results for $e_y$.
On the other hand, our method demonstrates superior control error standard deviations for $e_y$ and $e_\theta$, evidencing robustness.
Considering the success rate, the comprehensive evaluation shows that our method greatly outperforms the others.

Consequently, our method provides a guarantee for a high success rate by improving control precision on impact angle and lateral offset.
The Nonlinear approach struggles with the FoV constraints and is sensitive to noise.
Meanwhile, the MPC method, though better, often overreacts when converging to the goal pose, which is problematic within constrained spaces, leading to either a loss of sight or diminished precision.

\begin{figure*}[!tb]
    \centering
    \subfigure[]
        {\includegraphics[width=0.48\linewidth]{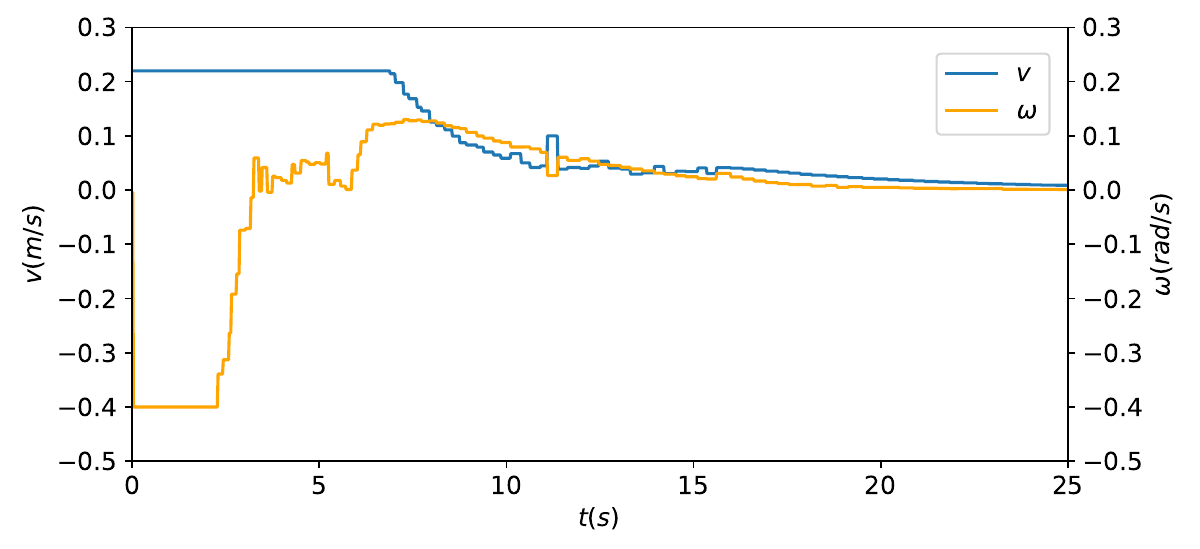}}
    \subfigure[]
        {\includegraphics[width=0.25\linewidth]{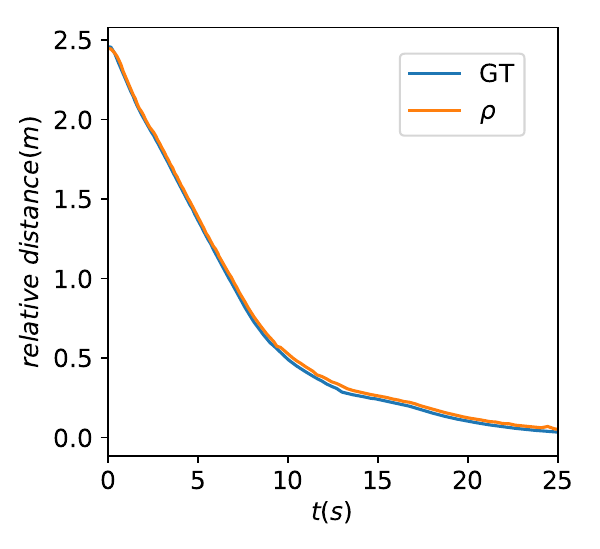}}
    \subfigure[]
        {\includegraphics[width=0.25\linewidth]{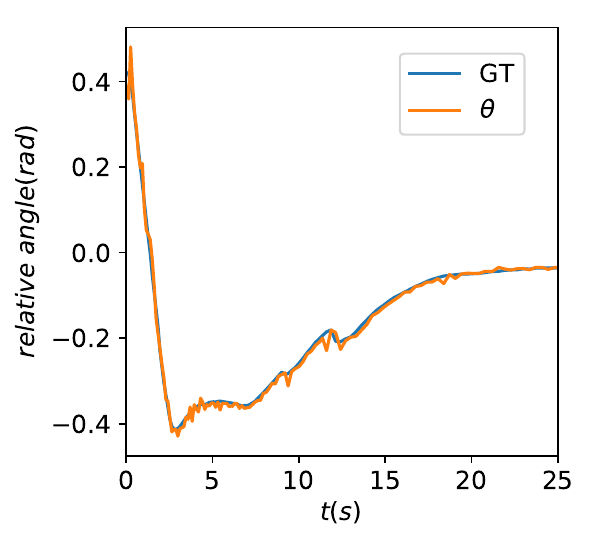}}\\
    \subfigure[]
        {\includegraphics[width=0.48\linewidth]{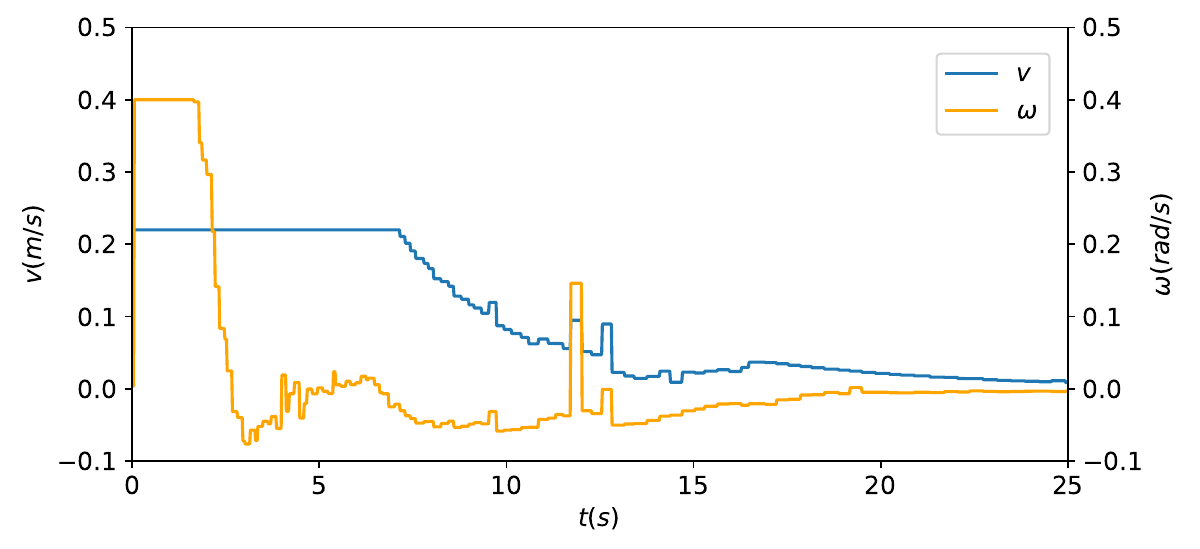}}
    \subfigure[]
        {\includegraphics[width=0.25\linewidth]{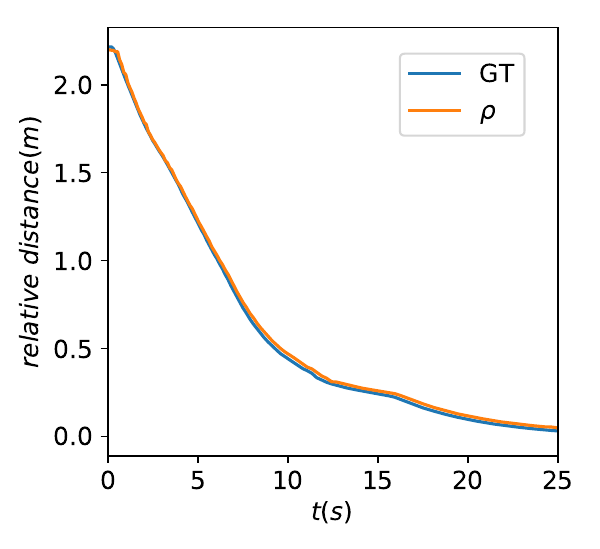}}
    \subfigure[]
        {\includegraphics[width=0.25\linewidth]{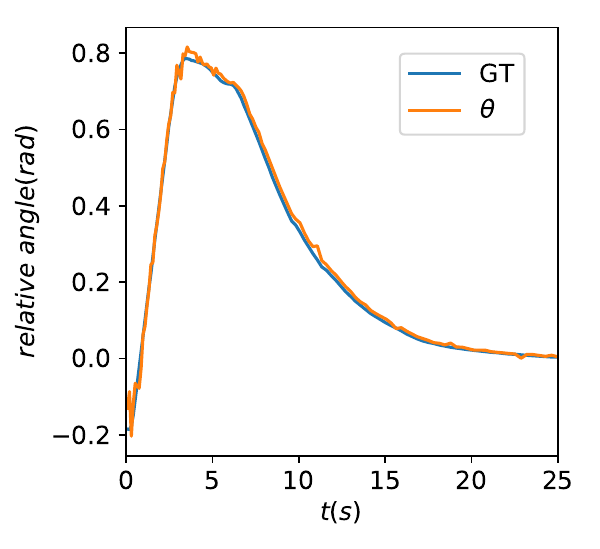}}
    \caption{
    The quantitative results of the two challenging control tasks. 
    (a) The linear velocity $v$ and angular velocity $\omega$ profile along the Trajectory 1.
    (b) Relative distance $\rho$ to the target trolley along the Trajectory 1. 
    (c) Relative orientation $\theta$ to the target trolley along the Trajectory 1. 
    (d) The linear velocity $v$ and angular velocity $\omega$ profile along the Trajectory 2.
    (e) Relative distance $\rho$ to the target trolley along the Trajectory 2. 
    (f) Relative orientation $\theta$ to the target trolley along the Trajectory 2. 
    }
    \label{fig:first_result_plot}
\end{figure*}

\subsection{Qualitative  Experiments}
Further detailed analysis shows the superiority of our approach in challenging scenarios.
\figref{fig:first_result_plot} exemplifies two such scenarios, where the trolley only partially appears on the side of the initial camera perspectives.
As the camera views from these two initial positions, the trolley partially appears on the side.
Our method can tackle this kind of safety-critical situation with a smooth trajectory and control, in contrast to the other methods which do not success.
The quantitative results are shown in \figref{fig:first_result_plot}, including the velocity profiles $v$ and $\omega$ and the relative observation of distance an angle $\rho $ and $ \theta$.
The robot achieves its goal in 21.35 s and 23.83 s for the respective trials as the velocity and errors converge to 0.
It first approaches the target at the maximum velocity and then adjusts the pose at a closer range.
Besides, comparing the observation and the ground truth in the two trajectories, the relative distance $\rho$ and angle $\theta$ between the robot and trolley are precisely perceived through our trolley localization method.
The experiments above demonstrate our robot’s impeccable approaching performance.

\subsection{Demonstration}

\begin{figure}[!tb]
    \centering
    {\includegraphics[width=0.98\linewidth]{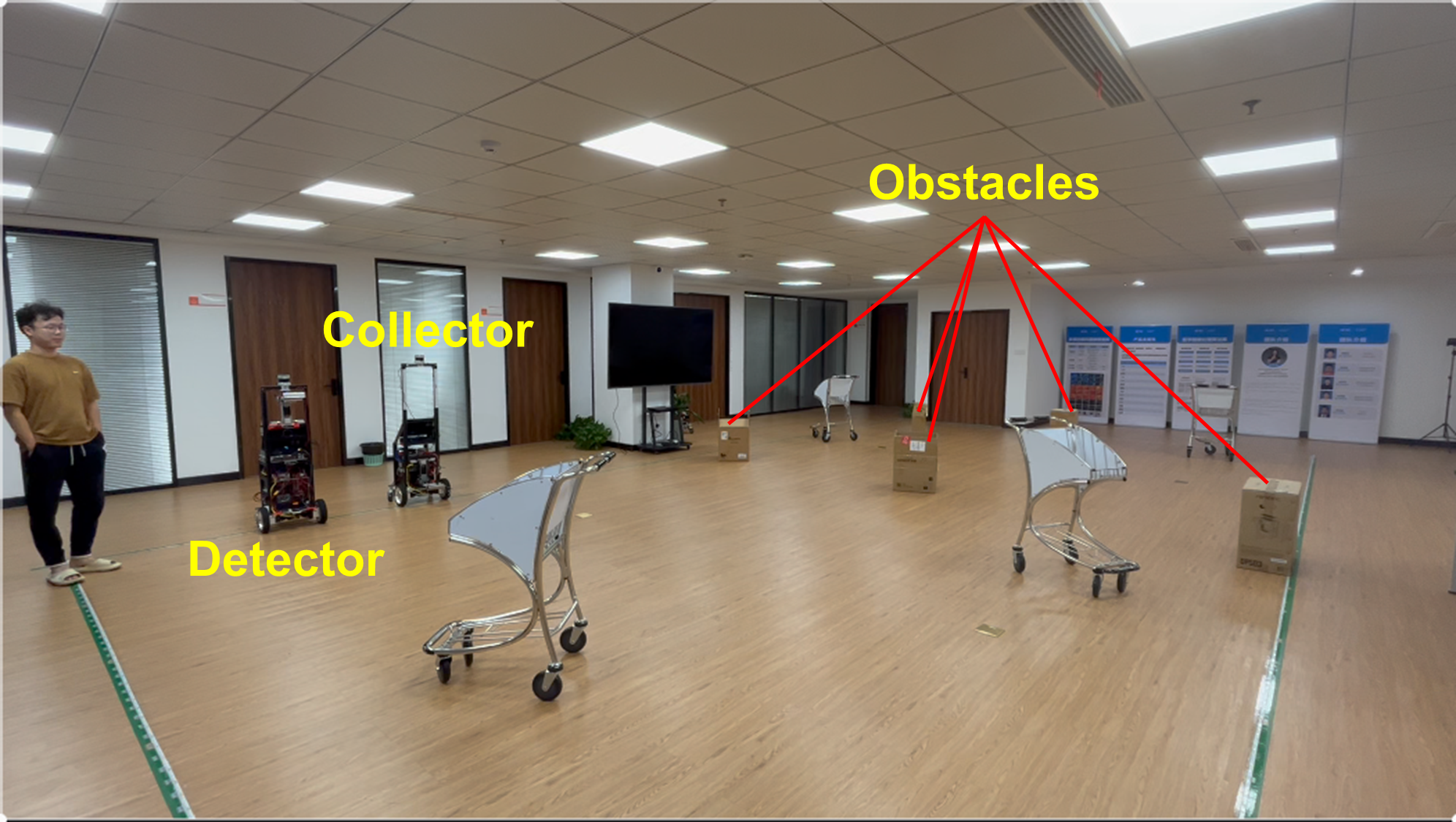}}
    \caption{
    The snapshot shows the setup of the demonstration scenario. The robots are assigned to collect the 4 randomly deposited trolleys avoiding pedestrian and unknown static obstacles.
    }
    \label{fig:environment_shot}
\end{figure}

We construct a real multi-trolley collection experiment as shown in \figref{fig:environment_shot} to verify the proposed robot system.
Within a $\text{20 m} \times \text{15 m}$ area, we establish a dynamic environment with randomly placed 5 unknown static obstacles and pedestrians.
The pedestrians randomly pass through except the Detector robot is near enough to the target and executing the Approaching or Docking stage. 
Prior to the experiment, we create the map shared between the Detector and Collector robots to facilitate global localization.
The experiment begin with the Detector and Collector robots positioned side-by-side in the lower-left corner of the testing area. 
Their primary objective is to identify and safely collect four scattered trolleys, arranging them into a queue.
Notably, the global localization of the trolley is not included.
We assume the collection sequence is the prior knowledge, and designate a series of waypoints as the transition of each collection trails.
The key parameters in the experiment are in Table \ref{tab:parameters}.

\begin{figure*}[tb!]
    \centering
    \subfigure[]
        {\includegraphics[width=0.48\linewidth]{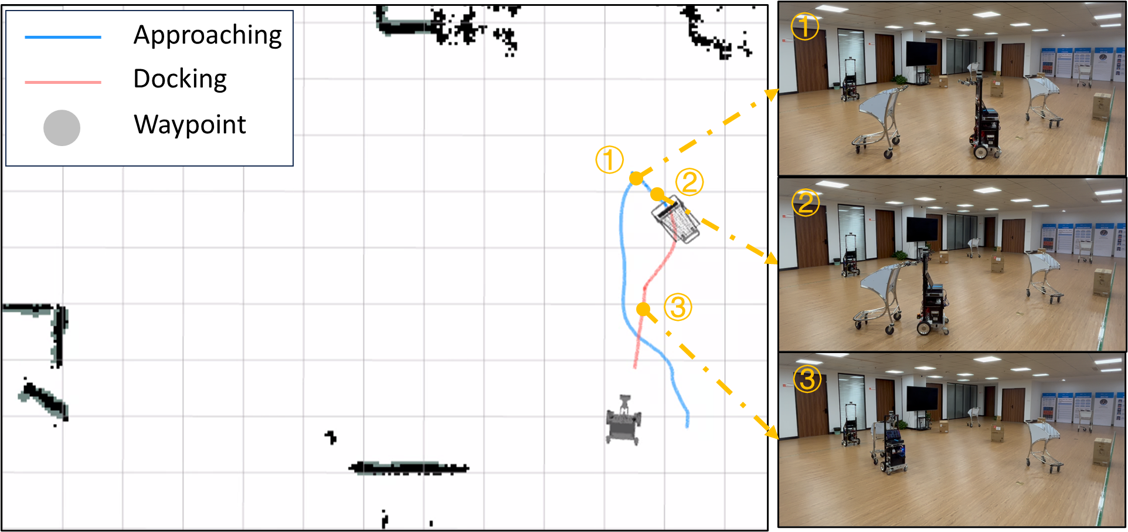}}
    \subfigure[]
        {\includegraphics[width=0.48\linewidth]{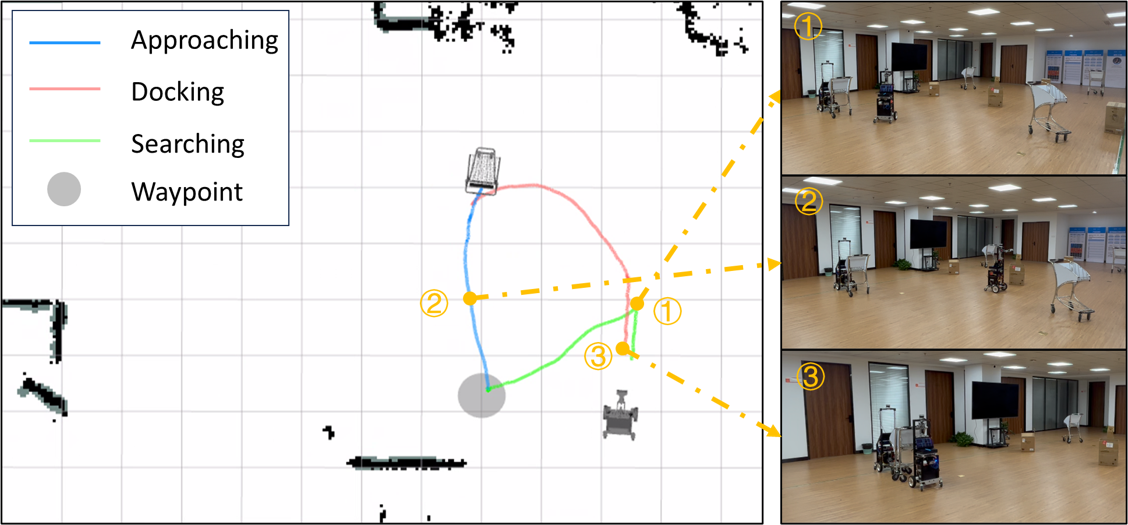}}\\
    \subfigure[]
        {\includegraphics[width=0.48\linewidth]{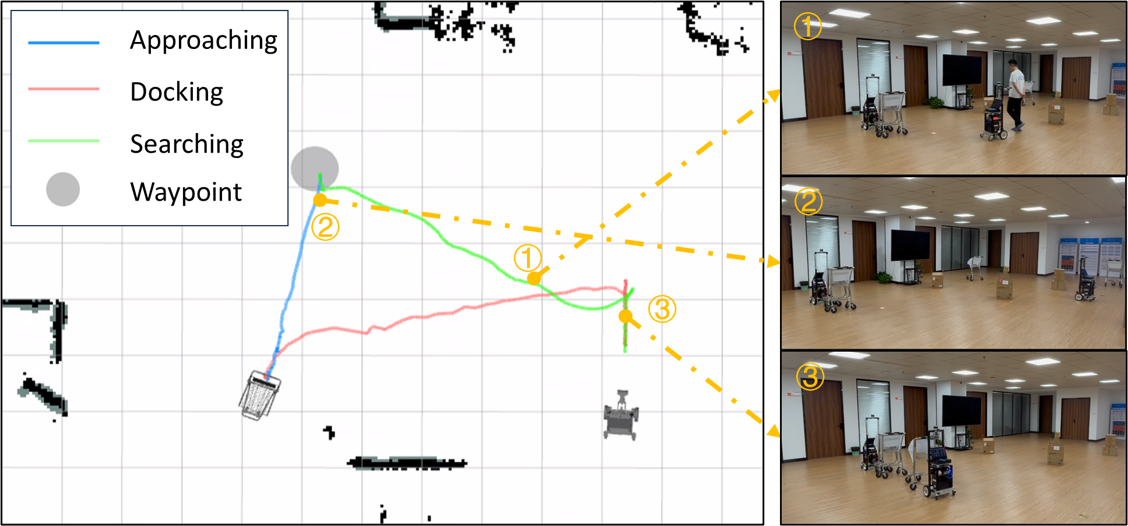}}
    \subfigure[]
        {\includegraphics[width=0.48\linewidth]{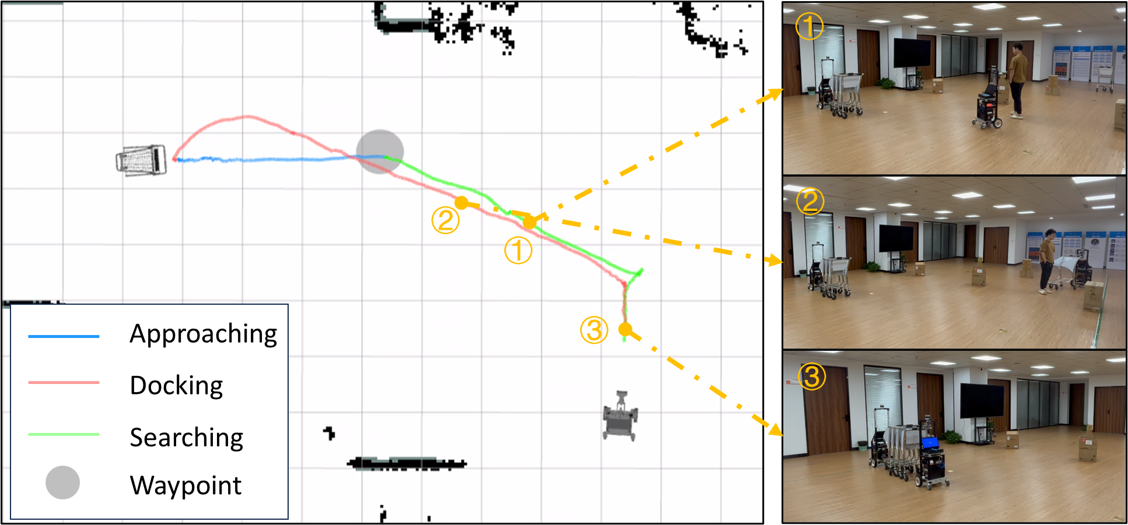}}
    \caption{
    Snapshots and visualization of the whole collection experiments in the real world. 
    The visualization shows the Collector robot's pose, the trolley's pose localized by the robot, and the Detector robot's trajectory in the Searching stage (green), the Approaching stage (blue), and the Docking stage (red).
    Each subfigure (a)-(d) shows a trail of collecting a trolley in order.
    }
    \label{fig:pipeline_experienment}
\end{figure*}

\begin{figure*}[tb!]
    \centering
    \subfigure[]
        {\includegraphics[width=0.48\linewidth]{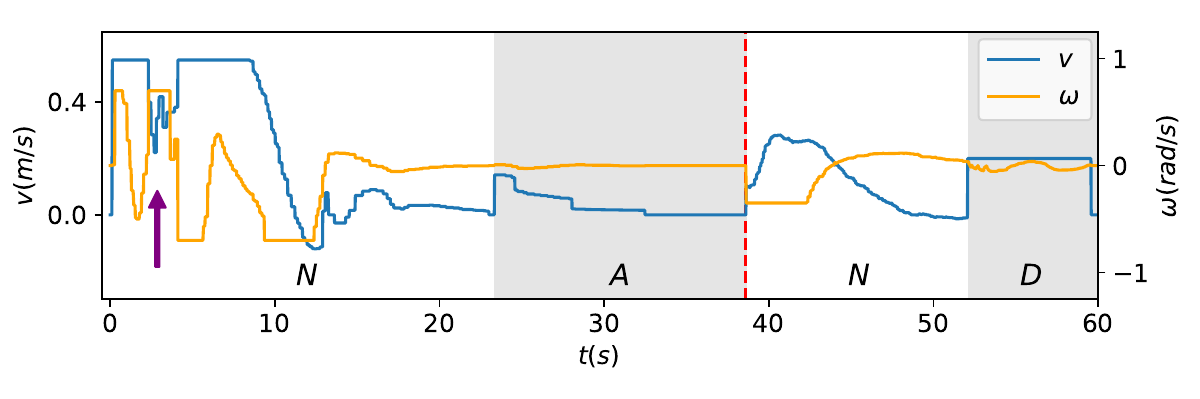}}
    \subfigure[]
        {\includegraphics[width=0.48\linewidth]{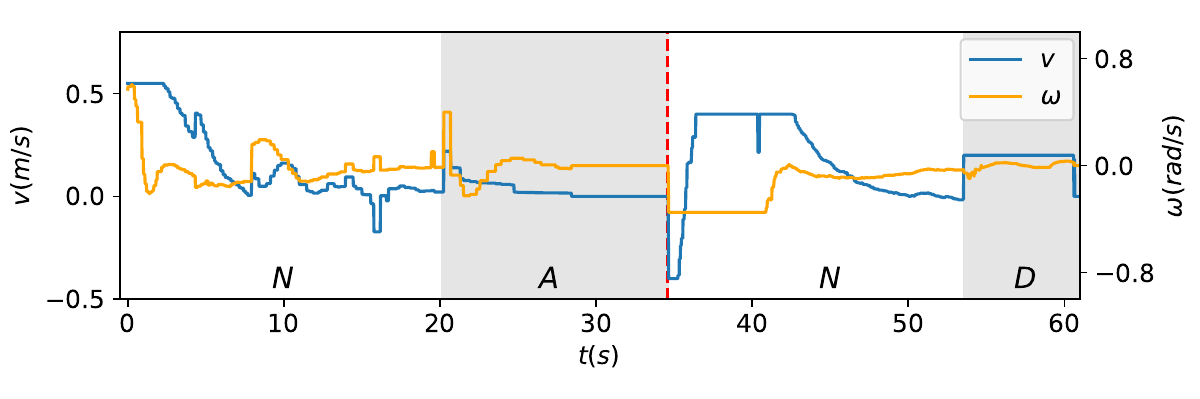}}
    \subfigure[]
        {\includegraphics[width=0.48\linewidth]{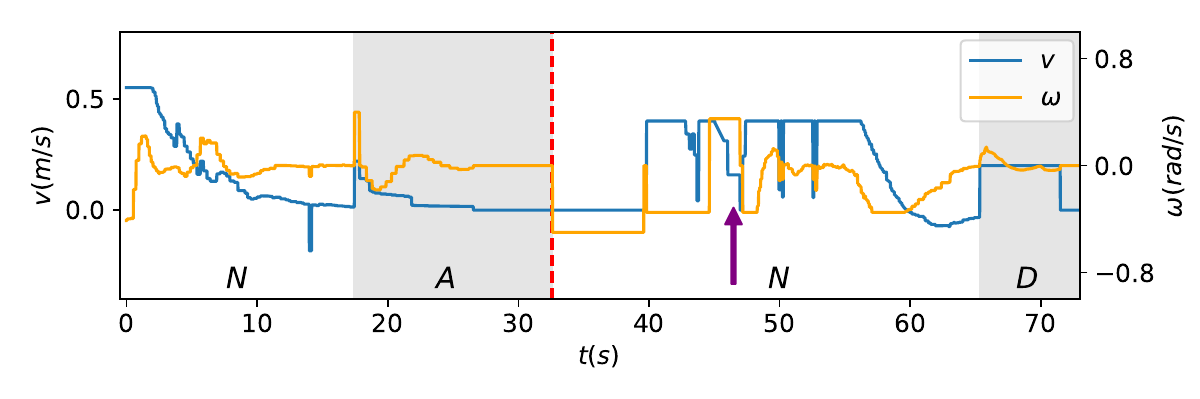}}
    \subfigure[]
        {\includegraphics[width=0.48\linewidth]{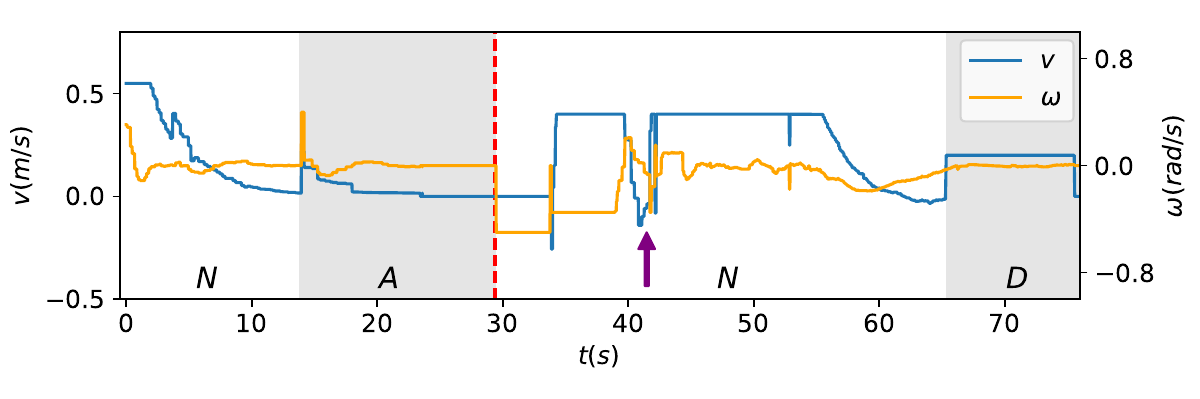}}
    \caption{
    Robot's linear velocity and angular velocity profile in the collection. 
    The shaded areas represent the Approaching (A) and Docking (D) stages, while the other is the Navigation stage (N). 
    The purple arrow indicates the robot is avoiding pedestrians.
    }
    \label{fig:pipeline_plot}
\end{figure*}

\figref{fig:pipeline_experienment} shows the whole collection process with the snapshots and visualization of the actual trajectory from SLAM.
The Collector robot remains in the initial pose during the whole process which is shown on the map.
The blue, red, and green trajectories respectively represent the Detector robot in different stages.
The Detector robot navigates to the waypoints (gray circles in \figref{fig:pipeline_experienment}) and localizes the trolley with onboard sensors after finishing a collection trail.

As a result, the experiment demonstrates that the proposed robotic system was capable of successfully and safely collecting all the deposited trolleys.
In \figref{fig:pipeline_experienment}(a), to collect the first trolley, the robot detects the trolley's orientation and navigates to its back.
Utilizing our vision-based control, the Detector robot accurately transports the first trolley to the target pose, and the Collector's manipulator catches the trolley, acting as the docking mechanism.
Then the Detector robot moves backward and navigates to the waypoint continuing the collection of the next trolley.
The collision avoidance of the passing pedestrian and the static obstacles is shown in \figref{fig:pipeline_experienment}(c) and \figref{fig:pipeline_experienment}(d).
The robot maneuvers safely with or without catching a trolley and plans the collision-free path according to the static obstacles' position.

Furthermore, \figref{fig:pipeline_plot} illustrates the Detector robot's velocity profiles except the Searching stage.
The red dashed lines depict the moments when the manipulator grasps the trolley, as the stages transition between the Approaching stage and the Navigation stage. 
The purple arrow indicates the velocity variation due to avoiding pedestrians.
For example, \figref{fig:pipeline_plot}(a) illustrates the data corresponding to the blue and red trajectory in \figref{fig:pipeline_experienment}(a).
Notably, between 2.3 s and 4.2 s, marked by purple arrows, the Detector robot reduces its linear velocity and adjusts its orientation to prevent a collision.
Another deceleration occurs at 9.6 s where the Detector robot replans its motion to slow down and even reverse because the trolley localization is corrected by the fine perception.
Then, at 23.3 s, the Detector robot is steered by the vision-based control, resulting in relatively small velocity changes.
After catching the trolley, the Detector robot navigates to the goal position which is queried from the Collector robot via UDP. 
Finally, the robot achieves docking by slightly adjusting orientation to align with the Detector robot and completes a collection in 59.6 s.
Similarly, the robot successfully completes the other 3 collections, which demonstrates the efficiency and robustness of our proposed system in complex and dynamic scenarios.


\section{Conclusions}
\label{sec:CON}

This paper introduces a novel autonomous robot system for the multiple-trolley collection task, paving the way for border application in airports.
We integrate an autonomous framework consisting of cost-effective hardware design, robust course-to-fine perception, dynamic environment motion planning, and optimization-base control.
The system employs a lightweight manipulator with the docking mechanism with limited degrees of freedom.
Our primary focus is a vision-based controller using the Control Lyapunov Function (CLF) and Control Barrier Function (CBF) within a Quadratic Programming (QP). 
This approach enhances tracking accuracy and field-of-view constraints, ensuring precise robot steering based on onboard sensors. 
Extensive experiments are conducted to evaluate the perception accuracy and compare the control method precision.
As a result, the proposed trolley localization method satisfies the requirement of high accuracy, and the CLF-CBF-QP controller shows outstanding performance in the trolley collection task.
Implemented in a real scenario, the robot safely and efficiently collects multiple trolleys showing the efficacy and reliability of the proposed framework and system.
Future work will focus on complicated multi-concat modeling in the collection problem with adjustable force control for more novel applications.





\section*{APPENDIX}
\begin{table}[htbp]
\caption{Cost of hardware for each robot}
\label{tab:cost_table}
\centering
    \begin{tabular}{cccccc}
    \toprule
    Modules & Motor & Structure &Battery & Sensors & MiniPC \\
    \midrule
    Price (\$) & 500 & 500 & 1000 & 5000 & 1000 \\
    \bottomrule
    \end{tabular}
\end{table}


\begin{table}[htbp]
\caption{EXPERIMENTAL PARAMETERS}
\label{tab:parameters}
\centering
\begin{tabular}{ccc}
\toprule
Symbol & Description &  Value \\

\midrule
$\xi$ &  \makecell{Offset of each collected trolley in Docking stage} & \text{0.18 m}
\\
$\rho_1$ & The Approaching stage triggers radius & \text{1.0 m}
\\
$\rho_2$ & The Docking stage triggers radius  & \text{2.1 m}
\\
$d_1$ & \makecell{Safety distance for robot without catching trolley} &  \text{0.55 m} 
\\
$d_{2}^f$ & \makecell{Forward safety distance for robot catching trolley} & \text{1.3 m}
\\
$d_{2}^b$ & \makecell{Backward safety distance for robot catching trolley} & \text{0.55 m} 
\\
 $v_{max}^n$ & Maximum linear velocity in Navigation stage  & \text{0.55 m/s}
\\
$\omega^n_{max}$ & Maximum angular velocity in Navigation stage  & \text{0.7 rad/s}
\\
$v_{max}^a$ & Maximum linear velocity in Approaching stage & \text{0.22 m/s}
\\
$\omega^a_{max}$ & \makecell{Maximum angular velocity in Approaching stage} & \text{0.4 rad/s} 
\\
$v_{max}^d$ & Maximum linear velocity in Docking stage  & \text{0.35 m/s}
\\
$\omega^d_{max}$ & Maximum angular velocity in Docking stage & \text{0.25 rad/s}
\\
\bottomrule
\end{tabular}
\end{table}




\bibliographystyle{IEEEtran}
\bibliography{root}

\begin{thebibliography}{10}
\providecommand{\url}[1]{#1}
\csname url@samestyle\endcsname
\providecommand{\newblock}{\relax}
\providecommand{\bibinfo}[2]{#2}
\providecommand{\BIBentrySTDinterwordspacing}{\spaceskip=0pt\relax}
\providecommand{\BIBentryALTinterwordstretchfactor}{4}
\providecommand{\BIBentryALTinterwordspacing}{\spaceskip=\fontdimen2\font plus
\BIBentryALTinterwordstretchfactor\fontdimen3\font minus
  \fontdimen4\font\relax}
\providecommand{\BIBforeignlanguage}[2]{{%
\expandafter\ifx\csname l@#1\endcsname\relax
\typeout{** WARNING: IEEEtran.bst: No hyphenation pattern has been}%
\typeout{** loaded for the language `#1'. Using the pattern for}%
\typeout{** the default language instead.}%
\else
\language=\csname l@#1\endcsname
\fi
#2}}
\providecommand{\BIBdecl}{\relax}
\BIBdecl

\bibitem{Hai2020}
W.~He, C.~Xue, X.~Yu, Z.~Li, and C.~Yang, ``Admittance-based controller design
  for physical human–robot interaction in the constrained task space,''
  \emph{IEEE Transactions on Automation Science and Engineering}, vol.~17,
  no.~4, pp. 1937--1949, 2020.

\bibitem{WANG2021100001}
J.~Wang, W.~Chen, X.~Xiao, Y.~Xu, C.~Li, X.~Jia, and M.~Q.-H. Meng, ``A survey
  of the development of biomimetic intelligence and robotics,''
  \emph{Biomimetic Intelligence and Robotics}, vol.~1, p. 100001, 2021.

\bibitem{teng2023fusionplanner}
S.~Teng, L.~Li, Y.~Li, X.~Hu, L.~Li, Y.~Ai, and L.~Chen, ``Fusionplanner: A
  multi-task motion planner for mining trucks using multi-sensor fusion
  method,'' \emph{arXiv preprint arXiv:2308.06931}, 2023.

\bibitem{Choi2019}
J.~Choi, K.~Park, M.~Kim, and S.~Seok, ``Deep reinforcement learning of
  navigation in a complex and crowded environment with a limited field of
  view,'' in \emph{2019 International Conference on Robotics and Automation
  (ICRA)}, 2019, pp. 5993--6000.

\bibitem{ye2023robot}
H.~Ye, J.~Zhao, Y.~Pan, W.~Cherr, L.~He, and H.~Zhang, ``Robot person following
  under partial occlusion,'' in \emph{2023 IEEE International Conference on
  Robotics and Automation (ICRA)}.\hskip 1em plus 0.5em minus 0.4em\relax IEEE,
  2023, pp. 7591--7597.

\bibitem{complex2022}
H.~Wang, L.~Zhang, Q.~Kong, W.~Zhu, J.~Zheng, L.~Zhuang, and X.~Xu, ``Motion
  planning in complex urban environments: An industrial application on
  autonomous last-mile delivery vehicles,'' \emph{Journal of Field Robotics},
  vol.~39, no.~8, pp. 1258--1285, 2022.

\bibitem{lowcost2021}
Y.~Du, B.~Mallajosyula, D.~Sun, J.~Chen, Z.~Zhao, M.~Rahman, M.~Quadir, and
  M.~K. Jawed, ``A low-cost robot with autonomous recharge and navigation for
  weed control in fields with narrow row spacing,'' in \emph{2021 IEEE/RSJ
  International Conference on Intelligent Robots and Systems (IROS)}, 2021, pp.
  3263--3270.

\bibitem{large-scale2021}
V.~Krátký, P.~Petráček, T.~Báča, and M.~Saska, ``An autonomous unmanned
  aerial vehicle system for fast exploration of large complex indoor
  environments,'' \emph{Journal of Field Robotics}, vol.~38, no.~8, pp.
  1036--1058, 2021.

\bibitem{xiao2022robotic}
A.~Xiao, H.~Luan, Z.~Zhao, Y.~Hong, J.~Zhao, W.~Chen, J.~Wang, and M.~Q.-H.
  Meng, ``Robotic autonomous trolley collection with progressive perception and
  nonlinear model predictive control,'' in \emph{2022 International Conference
  on Robotics and Automation (ICRA)}.\hskip 1em plus 0.5em minus 0.4em\relax
  IEEE, 2022, pp. 4480--4486.

\bibitem{wang2022DecisionMaking}
J.~Wang and M.~Q.-H. Meng, ``Real-time decision making and path planning for
  robotic autonomous luggage trolley collection at airports,'' \emph{IEEE
  Transactions on Systems, Man, and Cybernetics: Systems}, vol.~52, no.~4, pp.
  2174--2183, 2022.

\bibitem{xia2023collaborative}
B.~Xia, H.~Luan, Z.~Zhao, X.~Gao, P.~Xie, A.~Xiao, J.~Wang, and M.~Q.-H. Meng,
  ``Collaborative trolley transportation system with autonomous nonholonomic
  robots,'' in \emph{2023 IEEE/RSJ International Conference on Intelligent
  Robots and Systems (IROS)}, 2023, pp. 8046--8053.

\bibitem{pan2021}
J.~Pan, X.~Mai, C.~Wang, Z.~Min, J.~Wang, H.~Cheng, T.~Li, E.~Lyu, L.~Liu, and
  M.~Q.-H. Meng, ``A searching space constrained partial to full registration
  approach with applications in airport trolley deployment robot,'' \emph{IEEE
  Sensors Journal}, vol.~21, no.~10, pp. 11\,946--11\,960, 2021.

\bibitem{honerkamp2023n}
D.~Honerkamp, T.~Welschehold, and A.~Valada, ``N$^{2}$m$^{2}$: Learning
  navigation for arbitrary mobile manipulation motions in unseen and dynamic
  environments,'' \emph{IEEE Transactions on Robotics}, 2023.

\bibitem{Jauhri2022}
S.~Jauhri, J.~Peters, and G.~Chalvatzaki, ``Robot learning of mobile
  manipulation with reachability behavior priors,'' \emph{IEEE Robotics and
  Automation Letters}, vol.~7, no.~3, pp. 8399--8406, 2022.

\bibitem{yang2020human}
Y.~Yang, Y.~Pan, X.~Zhu, M.~Gao, J.~Zhang, and D.~Tao, ``A human-like
  dual-forklift collaborative mechanism for container handling,'' \emph{IEEE
  Transactions on Industrial Electronics}, vol.~68, no.~12, pp.
  12\,871--12\,880, 2020.

\bibitem{kashiri2019centauro}
N.~Kashiri, L.~Baccelliere, L.~Muratore, A.~Laurenzi, Z.~Ren, E.~M. Hoffman,
  M.~Kamedula, G.~F. Rigano, J.~Malzahn, S.~Cordasco \emph{et~al.}, ``Centauro:
  A hybrid locomotion and high power resilient manipulation platform,''
  \emph{IEEE Robotics and Automation Letters}, vol.~4, no.~2, pp. 1595--1602,
  2019.

\bibitem{vstibinger2021mobile}
P.~{\v{S}}tibinger, G.~Broughton, F.~Majer, Z.~Rozsyp{\'a}lek, A.~Wang,
  K.~Jindal, A.~Zhou, D.~Thakur, G.~Loianno, T.~Krajn{\'\i}k \emph{et~al.},
  ``Mobile manipulator for autonomous localization, grasping and precise
  placement of construction material in a semi-structured environment,''
  \emph{IEEE Robotics and Automation Letters}, vol.~6, no.~2, pp. 2595--2602,
  2021.

\bibitem{sustarevas2022autonomous}
J.~Sustarevas, D.~Kanoulas, and S.~Julier, ``Autonomous mobile 3d printing of
  large-scale trajectories,'' in \emph{2022 IEEE/RSJ International Conference
  on Intelligent Robots and Systems (IROS)}.\hskip 1em plus 0.5em minus
  0.4em\relax IEEE, 2022, pp. 6561--6568.

\bibitem{Bertoncelli2020}
F.~Bertoncelli, F.~Ruggiero, and L.~Sabattini, ``Linear time-varying mpc for
  nonprehensile object manipulation with a nonholonomic mobile robot,'' in
  \emph{2020 IEEE International Conference on Robotics and Automation (ICRA)},
  2020, pp. 11\,032--11\,038.

\bibitem{Tang2023}
Y.~Tang, H.~Zhu, S.~Potters, M.~Wisse, and W.~Pan, ``Unwieldy object delivery
  with nonholonomic mobile base: A stable pushing approach,'' \emph{IEEE
  Robotics and Automation Letters}, vol.~8, no.~11, pp. 7727--7734, 2023.

\bibitem{scholz2011cart}
J.~Scholz, S.~Chitta, B.~Marthi, and M.~Likhachev, ``Cart pushing with a mobile
  manipulation system: Towards navigation with moveable objects,'' in
  \emph{2011 IEEE International Conference on Robotics and Automation}.\hskip
  1em plus 0.5em minus 0.4em\relax IEEE, 2011, pp. 6115--6120.

\bibitem{schulze2023trajectory}
M.~Schulze, F.~Graaf, L.~Steffen, A.~Roennau, and R.~Dillmann, ``A trajectory
  planner for mobile robots steering non-holonomic wheelchairs in dynamic
  environments,'' in \emph{2023 IEEE International Conference on Robotics and
  Automation (ICRA)}.\hskip 1em plus 0.5em minus 0.4em\relax IEEE, 2023, pp.
  3642--3648.

\bibitem{burget2016bi}
F.~Burget, M.~Bennewitz, and W.~Burgard, ``Bi 2 rrt*: An efficient
  sampling-based path planning framework for task-constrained mobile
  manipulation,'' in \emph{2016 IEEE/RSJ International Conference on
  Intelligent Robots and Systems (IROS)}.\hskip 1em plus 0.5em minus
  0.4em\relax IEEE, 2016, pp. 3714--3721.

\bibitem{Aguilera2023}
S.~Aguilera, M.~A. Murtaza, J.~Rogers, and S.~Hutchinson, ``Modeling and
  inertial parameter estimation of cart-like nonholonomic systems using a
  mobile manipulator,'' in \emph{2023 IEEE International Conference on Robotics
  and Automation (ICRA)}, 2023, pp. 3073--3079.

\bibitem{mathew2015multirobot}
N.~Mathew, S.~L. Smith, and S.~L. Waslander, ``Multirobot rendezvous planning
  for recharging in persistent tasks,'' \emph{IEEE Transactions on Robotics},
  vol.~31, no.~1, pp. 128--142, 2015.

\bibitem{narvaez2020autonomous}
E.~Narv{\'a}ez, A.~A. Ravankar, A.~Ravankar, T.~Emaru, and Y.~Kobayashi,
  ``Autonomous vtol-uav docking system for heterogeneous multirobot team,''
  \emph{IEEE Transactions on Instrumentation and Measurement}, vol.~70, pp.
  1--18, 2020.

\bibitem{he2023image}
G.~He, Y.~Jangir, J.~Geng, M.~Mousaei, D.~Bai, and S.~Scherer, ``Image-based
  visual servo control for aerial manipulation using a fully-actuated uav,''
  \emph{arXiv preprint arXiv:2306.16530}, 2023.

\bibitem{xing2023autonomous}
J.~Xing, G.~Cioffi, J.~Hidalgo-Carri{\'o}, and D.~Scaramuzza, ``Autonomous
  power line inspection with drones via perception-aware mpc,'' \emph{arXiv
  preprint arXiv:2304.00959}, 2023.

\bibitem{chen2021virtual}
Y.~Chen, D.~F. Paez-Granados, B.~Leme, and K.~Suzuki, ``Virtual landmark-based
  control of docking support for assistive mobility devices,'' \emph{IEEE/ASME
  Transactions on Mechatronics}, vol.~26, no.~4, pp. 2007--2015, 2021.

\bibitem{pankert2020perceptive}
J.~Pankert and M.~Hutter, ``Perceptive model predictive control for continuous
  mobile manipulation,'' \emph{IEEE Robotics and Automation Letters}, vol.~5,
  no.~4, pp. 6177--6184, 2020.

\bibitem{peric2021direct}
L.~Peric, M.~Brunner, K.~Bodie, M.~Tognon, and R.~Siegwart, ``Direct force and
  pose nmpc with multiple interaction modes for aerial push-and-slide
  operations,'' in \emph{2021 IEEE International Conference on Robotics and
  Automation (ICRA)}.\hskip 1em plus 0.5em minus 0.4em\relax IEEE, 2021, pp.
  131--137.

\bibitem{ames2016control}
A.~D. Ames, X.~Xu, J.~W. Grizzle, and P.~Tabuada, ``Control barrier function
  based quadratic programs for safety critical systems,'' \emph{IEEE
  Transactions on Automatic Control}, vol.~62, no.~8, pp. 3861--3876, 2016.

\bibitem{jankovic2018robust}
M.~Jankovic, ``Robust control barrier functions for constrained stabilization
  of nonlinear systems,'' \emph{Automatica}, vol.~96, pp. 359--367, 2018.

\bibitem{wang2017safety}
L.~Wang, A.~D. Ames, and M.~Egerstedt, ``Safety barrier certificates for
  collisions-free multirobot systems,'' \emph{IEEE Transactions on Robotics},
  vol.~33, no.~3, pp. 661--674, 2017.

\bibitem{bena2023safety}
R.~M. Bena, C.~Zhao, and Q.~Nguyen, ``Safety-aware perception for autonomous
  collision avoidance in dynamic environments,'' \emph{IEEE Robotics and
  Automation Letters}, 2023.

\bibitem{ames2019control}
A.~D. Ames, S.~Coogan, M.~Egerstedt, G.~Notomista, K.~Sreenath, and P.~Tabuada,
  ``Control barrier functions: Theory and applications,'' in \emph{2019 18th
  European control conference (ECC)}.\hskip 1em plus 0.5em minus 0.4em\relax
  IEEE, 2019, pp. 3420--3431.

\bibitem{andersson2019casadi}
J.~A. Andersson, J.~Gillis, G.~Horn, J.~B. Rawlings, and M.~Diehl,
  ``{{CasADi}}: A software framework for nonlinear optimization and optimal
  control,'' \emph{Mathematical Programming Computation}, vol.~11, no.~1, pp.
  1--36, 2019.

\bibitem{biegler2009large}
L.~T. Biegler and V.~M. Zavala, ``Large-scale nonlinear programming using
  {{IPOPT}}: An integrating framework for enterprise-wide dynamic
  optimization,'' \emph{Computers \& Chemical Engineering}, vol.~33, no.~3, pp.
  575--582, 2009.

\bibitem{yolov5}
\BIBentryALTinterwordspacing
G.~R. Jocher, A.~Stoken, J.~Borovec, NanoCode, A.~Chaurasia, TaoXie,
  L.~Changyu, Abhiram, Laughing, tkianai, yxNONG, A.~Hogan, lorenzomammana,
  AlexWang, J.~H{\'a}jek, L.~Diaconu, Marc, Y.~Kwon, Oleg, wanghaoyang,
  Y.~Defretin, A.~Lohia, ml~ah, B.~Milanko, B.~Fineran, D.~P. Khromov,
  D.~Yiwei, Doug, Durgesh, and F.~Ingham, ``ultralytics/yolov5: v5.0 -
  yolov5-p6 1280 models, aws, supervise.ly and youtube integrations,'' 2021.
  [Online]. Available: \url{https://api.semanticscholar.org/CorpusID:244964519}
\BIBentrySTDinterwordspacing

\bibitem{hourglass}
\BIBentryALTinterwordspacing
A.~Newell, K.~Yang, and J.~Deng, ``Stacked hourglass networks for human pose
  estimation,'' in \emph{European Conference on Computer Vision}, 2016.
  [Online]. Available: \url{https://api.semanticscholar.org/CorpusID:13613792}
\BIBentrySTDinterwordspacing

\bibitem{fastlio}
W.~Xu, Y.~Cai, D.~He, J.~Lin, and F.~Zhang, ``Fast-lio2: Fast direct
  lidar-inertial odometry,'' \emph{IEEE Transactions on Robotics}, vol.~38,
  no.~4, pp. 2053--2073, 2022.

\bibitem{gicp}
\BIBentryALTinterwordspacing
A.~V. Segal, D.~H{\"a}hnel, and S.~Thrun, ``Generalized-icp,'' in
  \emph{Robotics: Science and Systems}, 2009. [Online]. Available:
  \url{https://api.semanticscholar.org/CorpusID:231748613}
\BIBentrySTDinterwordspacing

\bibitem{octomap}
\BIBentryALTinterwordspacing
A.~Hornung, K.~M. Wurm, M.~Bennewitz, C.~Stachniss, and W.~Burgard,
  ``{OctoMap}: An efficient probabilistic {3D} mapping framework based on
  octrees,'' \emph{Autonomous Robots}, 2013, software available at
  \url{https://octomap.github.io}. [Online]. Available:
  \url{https://octomap.github.io}
\BIBentrySTDinterwordspacing

\end{thebibliography}

\end{document}